\documentclass[table]{article}

\usepackage{fontawesome}
\usepackage[numbers, compress]{natbib}

\usepackage{microtype}
\usepackage{graphicx}
\usepackage{subfigure}
\usepackage{booktabs} %
\usepackage{multirow}
\usepackage{tcolorbox}
\usepackage{dashrule} %
\usepackage{wrapfig}

\usepackage{xcolor}

\usepackage{hyperref}
\usepackage{enumitem}
\setlist[itemize]{leftmargin=*}
\setlist[enumerate]{leftmargin=*}

\newif\ificlrtemplate
\iclrtemplatefalse

\usepackage[accepted]{icml2025_arxiv}

\usepackage{amsmath}
\usepackage{amssymb}
\usepackage{mathtools}
\usepackage{amsthm}
\usepackage{nicefrac}

\usepackage[capitalize,noabbrev]{cleveref}

\theoremstyle{plain}

\theoremstyle{definition}

\theoremstyle{remark}

\usepackage[textsize=tiny]{todonotes}

\usepackage{xspace}

\definecolor{lasallegreen}{rgb}{0.03, 0.47, 0.19}
\definecolor{junglegreen}{rgb}{0.16, 0.67, 0.53}
\definecolor{hotmagenta}{rgb}{1.0, 0.11, 0.81}

\newcommand{\dashline}{\hdashrule[0.5ex]{\linewidth}{0.5pt}{3pt}}

\newcommand{\SAME}{\textsf{\textcolor{lasallegreen}{MATH-P-Simple}}\xspace}
\newcommand{\HARD}{\textsf{\textcolor{brown}{MATH-P-Hard}}\xspace}
\newcommand{\Original}{\textsf{Original}\xspace}
\usepackage[scaled=0.98]{inconsolata}
\usepackage[T1]{fontenc}
\usepackage[scaled=0.98]{biolinum}

\newcommand{\red}[1]{#1}
\newcommand{\after}[1]{#1}

\icmltitlerunning{MATH-Perturb: Benchmarking LLMs' Math Reasoning Abilities against Hard Perturbations}

\begin{document}

\twocolumn[
\icmltitle{MATH-Perturb: Benchmarking LLMs' Math Reasoning Abilities \\ against Hard Perturbations}

\icmlsetsymbol{equal}{$\dagger$}

\begin{icmlauthorlist}

\icmlauthor{Kaixuan Huang}{princeton}
\icmlauthor{Jiacheng Guo}{equal,princeton}
\icmlauthor{Zihao Li}{equal,princeton}
\icmlauthor{Xiang Ji}{equal,princeton}
\icmlauthor{Jiawei Ge}{equal,princeton}
\icmlauthor{Wenzhe Li}{equal,princeton}
\icmlauthor{Yingqing Guo}{equal,princeton}
\icmlauthor{Tianle Cai}{equal,princeton}
\icmlauthor{Hui Yuan}{equal,princeton}
\icmlauthor{Runzhe Wang}{equal,princeton}
\icmlauthor{Yue Wu}{equal,princeton}
\icmlauthor{Ming Yin}{equal,princeton}
\icmlauthor{Shange Tang}{equal,princeton}
\\
\icmlauthor{Yangsibo Huang}{google}
\icmlauthor{Chi Jin}{princeton}
\icmlauthor{Xinyun Chen}{google}
\icmlauthor{Chiyuan Zhang}{google}
\icmlauthor{Mengdi Wang}{princeton}
\end{icmlauthorlist}

\icmlaffiliation{princeton}{Princeton University}
\icmlaffiliation{google}{Google}

\icmlcorrespondingauthor{Kaixuan Huang}{kaixuanh@princeton.edu}

\icmlkeywords{Machine Learning, LLM}

\vskip 0.3in
]

\printAffiliationsAndNotice{\icmlEqualContribution} %

\setlength{\textfloatsep}{4pt}

\begin{abstract}
    Large language models have demonstrated impressive performance on challenging mathematical reasoning tasks, which has triggered the discussion of whether the performance is achieved by true reasoning capability or memorization. 
To investigate this question, prior work has constructed mathematical benchmarks when questions undergo \textit{simple perturbations} -- modifications that still preserve the underlying reasoning patterns of the solutions. However, no work has explored \textit{hard perturbations}, which fundamentally change the nature of the problem so that the original solution steps do not apply. To bridge the gap, we construct \SAME and \HARD via simple perturbation and hard perturbation, respectively. Each consists of 279 perturbed math problems derived from level-5 (hardest) problems in the MATH dataset~\citep{hendrycksmath2021}. We observe significant performance drops on \HARD across various models, including o1-mini {($-16.49$\%)} and gemini-2.0-flash-thinking ($-12.9$\%).
We also raise concerns about a novel form of memorization where models blindly apply learned problem-solving skills without assessing their applicability to modified contexts. This issue is amplified when using original problems for in-context learning. 
We call for research efforts to address this challenge, which is critical for developing more robust and reliable reasoning models.

\end{abstract}

\section{Introduction}

    \begin{figure*}[htbp]
    \centering
        \vspace{-1em}
            \begin{minipage}[c]{0.31\textwidth}  
                \centering
                 \begin{minipage}[b]{\textwidth}  
                    \centering
                        {\textcolor{gray}{\textbf{Overview of MATH-Perturb Benchmark}}}
                    \vspace{0.8em}
                \end{minipage}
                 \begin{minipage}[b]{0.9\textwidth}  
                     \centering
                     \includegraphics[width=\textwidth]{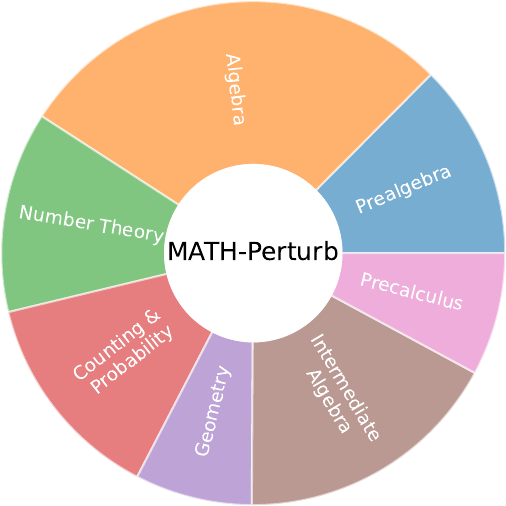}    
                     \vspace{0.05em}
                 \end{minipage}
                \begin{minipage}[b]{\textwidth}  
                    \centering
                    \resizebox{0.95\textwidth}{!}{
                    \begin{tabular}{l|c|c}
                        \toprule
                        \textbf{Split} & \textbf{Type} & \textbf{Size}   \\ \midrule
                        \SAME   & Simple Perturbation & 279    \\ \midrule
                        \HARD   & Hard Perturbation& 279   \\ \bottomrule
                    \end{tabular} 
                    }
                \end{minipage}
            \end{minipage}
        \hfill
        \begin{minipage}[c]{0.68\textwidth}  
            \centering
            \includegraphics[width=\textwidth]{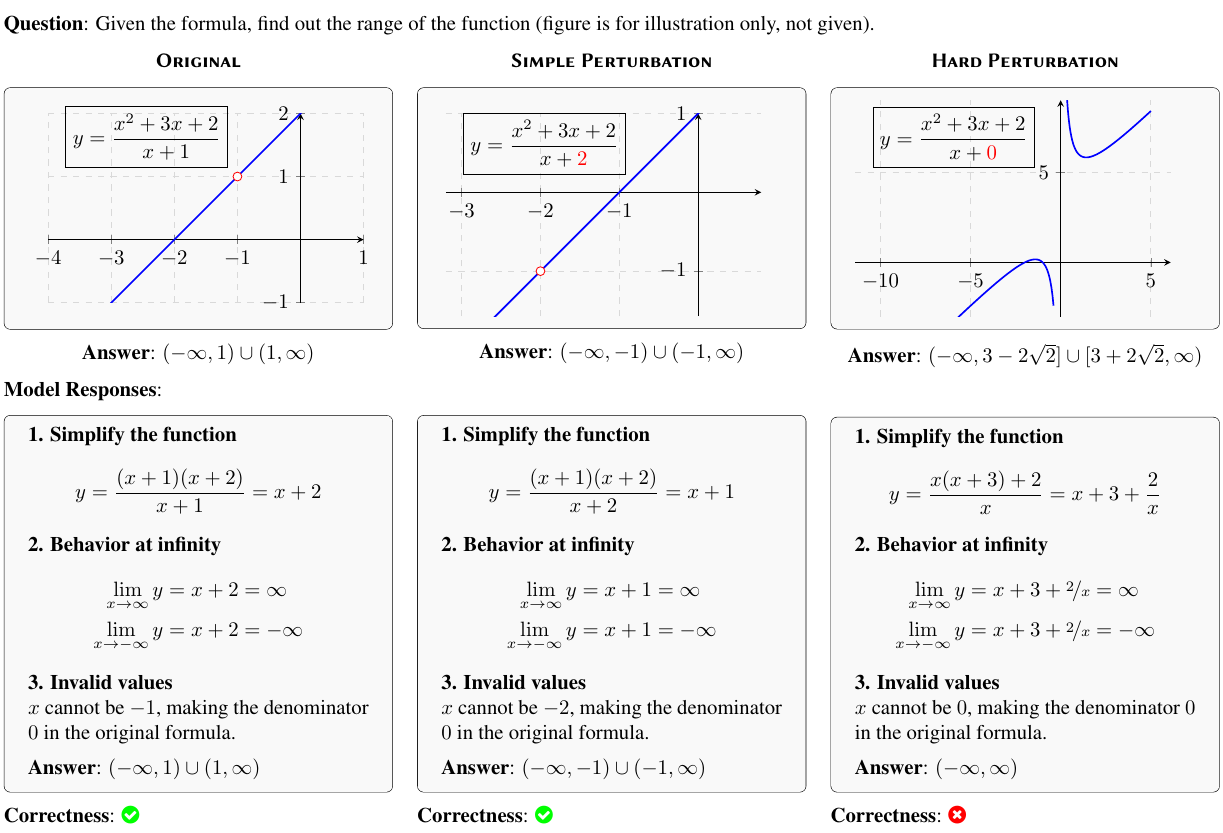}
        \end{minipage}
        \vspace{-1em}
    \caption{ \textbf{Left:} The overview of MATH-Perturb Benchmark. \textbf{Right:} An example of the original problem, its \textbf{simple} perturbation, its \textbf{hard} perturbation, and the corresponding model responses that \red{overfit} the short-cut solution. The simple perturbation to the problem is non-essential, so the modified problem can be solved using the same method as the original problem. The hard perturbation changes the problem fundamentally and it requires more difficult problem-solving skills. The shortcut solution can solve the original problem and its simple perturbation but fails on the hard perturbation.  } %
    \label{fig:hard_perturb}
    \end{figure*}

Large language models (LLMs) have achieved remarkable progress in solving many previously challenging tasks and demonstrating signs of general intelligence~\citep{bubeck2023sparks}. As LLMs become more intelligent, the research community responds by developing and adopting new benchmarks to guide the development of better  models~\citep{wang2024mmlu, zhou2023instruction, liu2024your, rein2023gpqa, berkeley-function-calling-leaderboard}. 

In mathematical reasoning, the field has progressed from simpler datasets like SVAMP~\citep{patel-etal-2021-nlp} and GSM8K~\citep{cobbe2021gsm8k} to more challenging benchmarks such as MATH~\citep{hendrycksmath2021}, OlympiadBench~\citep{he2024olympiadbench}, and AIME problems. Models continue to strike higher performance on these advanced benchmarks through stronger architectures, novel training approaches, and better training data~\citep{openaio1, yang2024qwen25, shao2024deepseekmath, deepseekai2025deepseekr1incentivizingreasoningcapability}. \nocite{qwq-32b-preview}

Nevertheless, concerns about data contamination and out-of-distribution generalization remain. Model performance can be artificially high if variants of the evaluation set leak into the training datasets or if its distribution is over-represented. In these cases, the model could be merely doing pattern recognition and memorizing the solution steps without understanding the underlying rationale, making it vulnerable to perturbations of the problem formulation~\citep{zhang2024careful, srivastava2024functional}. %

Several works have been proposed to quantify the robustness of reasoning models against such perturbations~\citep{shi2023large, mirzadeh2024gsm, zhang2024careful, srivastava2024functional, gulati2024putnamaxiom, zou2024dynamath}. Notably, \citet{srivastava2024functional} created Functional-MATH by manually rewriting the original problems in the MATH benchmark~\citep{hendrycksmath2021}  into problem templates, where the numerical values in the problem statements and the corresponding answers can be varied automatically to generate infinitely-many versions that use the same math problem-solving skills. They observed performance drops between the modified benchmark and the original benchmark for several state-of-the-art language models, indicating that those models are indeed \textit{biased} towards the original configurations of numerical values due to some form of data contamination. However, most existing work focuses on perturbing non-critical parameters (e.g., numerical values) that do not alter the fundamental reasoning patterns required to solve the problem. We refer to such changes as \textbf{simple perturbations}. While prior studies have shown that LLMs can generalize across a range of problem variants by relying on bag-of-heuristics reasoning~\citep{nikankin2024arithmetic,OthelloGPT}, this form of generalization does not necessarily reflect a true understanding of the underlying principles. As a result, models may still fail when faced with a substantial shift in reasoning patterns.

In this work, we take one step forward beyond simple perturbations. We consider \textbf{hard perturbations}: while at lexical level (e.g. edit distance) the modification is similar to simple perturbations, we ensure to change the problem formulations fundamentally so that the original solution paths are no longer applicable to the perturbed settings; see \cref{fig:hard_perturb} for a comparison between the two types of perturbations. 
A genuinely robust reasoning model that understands the underlying rationales should \textit{not only} solve the modified problems under simple perturbations \textit{but also} be able to judge whether the problem formulations change in a way that fundamentally alters the problems and respond accordingly, instead of applying the learned skills indiscriminately.

As the capabilities of large language models continue to advance and the average-case performance continue to improve, the \after{generalization abilities against} hard perturbations may soon become the primary bottleneck in their real-world usages. Addressing this challenge will be critical for advancing the robustness and reliability of future LLMs.

We summarize our contributions and key findings below:
\begin{itemize}[itemsep=1pt, parsep=1pt, topsep=1pt]
    \item We design and construct \SAME (simple perturbation) and \HARD (hard perturbation), each consisting of 279 perturbed math problems that originate from the level-5 (hardest) problems of the MATH dataset~\citep{hendrycksmath2021}.  The datasets are curated by 12 graduate-level experts with rigorous rubrics and cross-checking workflow for quality control (\cref{sec:data-curation}).
    \item We benchmark the math reasoning abilities of 18 LLMs (\cref{sec:benchmarking}), and show that all the models, including o1-mini and gemini-2.0-flash-thinking, suffer significant performance drops {(10\%-25\%)} on \HARD. This indicates these models are biased towards the original distribution of reasoning patterns and suffer from \textit{out-of-distribution effect} when facing problems with hard perturbations. 
    \item We conduct in-depth failure mode analysis (\cref{sec:failure:mode}) and identify a new form of memorization, where the model memorizes the problem-solving techniques from the training set and blindly applies them without judging whether the modified settings are still suitable.
    \item We investigate the influences of in-context learning (ICL) with the corresponding original unmodified problem and solution (\cref{sec:icl}), and demonstrate that ICL with original example may hurt the model on \HARD, as the models may fail to recognize the subtle differences and get misled by the demonstration.
\end{itemize}

    \begin{figure*}[ht]
    \centering
        \includegraphics[width=0.98\linewidth]{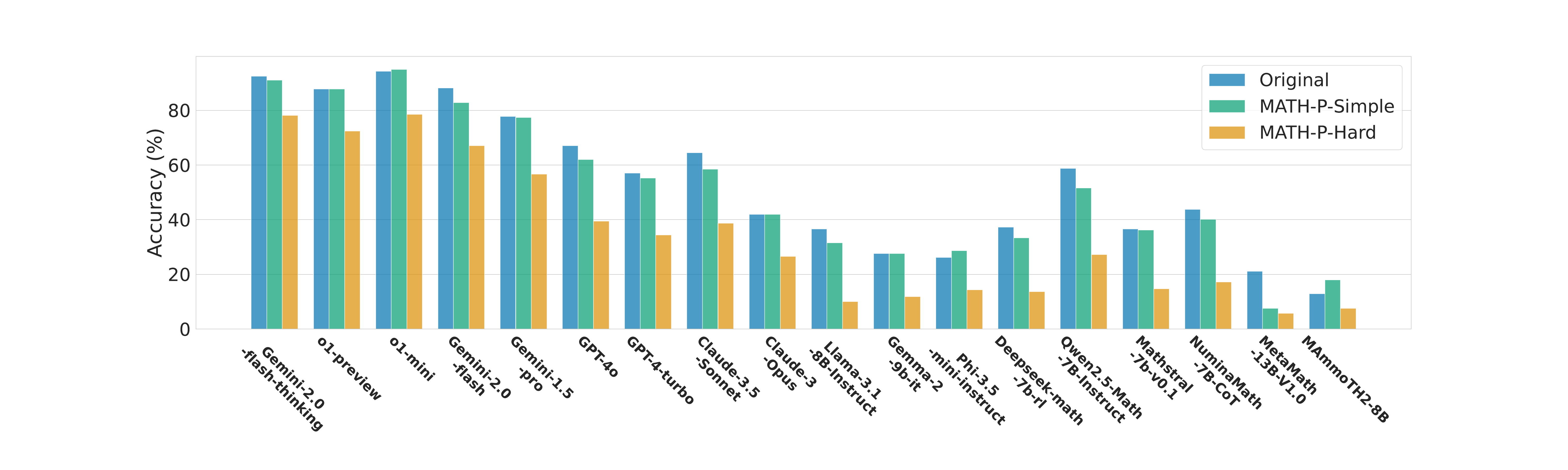}
        \vspace{-5mm}
    \caption{Performance on \SAME, \HARD, and the corresponding \Original problems. We observe performance degradations across all models on \HARD. 
    }
    \label{fig:performance}
    \end{figure*}

\section{Dataset Curation}
\label{sec:data-curation}

\begin{figure*}[t]
\centering
    \includegraphics[width=0.95\linewidth]{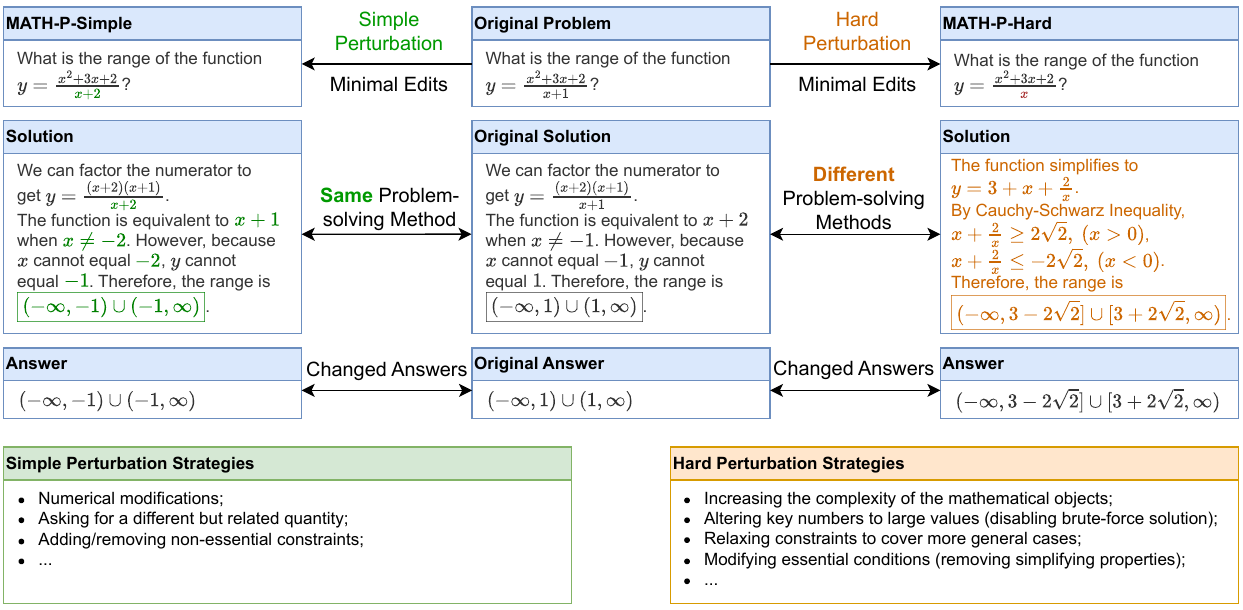}
\caption{Illustration of the annotation process for \SAME and \HARD. }
\label{fig:annotation}
\end{figure*}

\textbf{Origin of the Dataset.} 
We choose the popular MATH benchmark~\citep{hendrycksmath2021}, which contains challenging mathematical reasoning problems sourced from American high school mathematics competitions such as the AMC 10, AMC 12, and AIME. Each problem belongs to one of the 7 subjects: Prealgebra, Algebra, Number Theory, Counting and Probability, Geometry, Intermediate Algebra, and Precalculus. Besides, each problem is labeled with a difficulty level of 1 (easiest) to 5 (hardest). The problems may contain LaTeX and Asymptote graphics language for describing mathematical concepts and geometric figures.

As the state-of-the-art reasoning models can already solve MATH problems with overall accuracies higher than 90\%~\citep{openaio1, team2024gemini, deepseekai2025deepseekr1incentivizingreasoningcapability}, we opt to focus only on the hardest level-5 problems in our work, and create new benchmarks from these level-5 problems.
We use level-5 problems from both the \texttt{train} split and the \texttt{test} split as the seed problems, so we are able to investigate whether language models behave differently on the two splits.

\textbf{Annotation Criterion.}
For each problem, we modify the problem to create two variations:

(1) for \SAME, we make \textbf{simple perturbations}, i.e., non-essential modifications to the problem, ensuring that the modified problem can be solved using \textit{the same method} as the original problem. 

(2) for \HARD, we make \textbf{hard perturbations}, i.e., small but fundamental modifications to the problem so that the modified problem \emph{cannot} be solved using \textit{the same method} as the original problem. Instead, it requires deeper math understanding and harder problem-solving skills.

Besides, we ensure the following two additional requirements:

\begin{itemize}[itemsep=1pt, parsep=1pt, topsep=1pt]
\item \textbf{Minimal Edits:} To test the \after{generalization} of the reasoning models and elicit potential memorization behaviors, we ask the annotators to make as minimal modifications as possible. Therefore, the modified problems stay close to the original problems in the text form.  

\item \textbf{Changed Answers:} For both of the modifications, we guarantee that the answers to the modified problems are different from the original answer. Therefore, models cannot cheat by pattern recognition and outputting memorized solutions. %
\end{itemize}

\textbf{Quality Control.}
We recruited 12 annotators (PhD students) with strong mathematical backgrounds for the annotation task. All the annotators hold a bachelor's degree in mathematics, have done researches in theoretical machine learning, and/or competed in mathematical competitions during high school.

To ensure the quality of the benchmark, all the annotators are required to double-check their annotations. Each modified problem is also cross-validated by an independent annotator to make sure the answer is correct.

Additionally, we manually went through all the problems where the o1-mini's answer and the annotated answer differ and confirmed that the annotated answers are correct. %

\textbf{Benchmark Overview and Statistics.}

After removing several annotations that failed the quality checks, we obtained 279 pairs of modifications, where 164 examples are from \texttt{train} split and 115 examples are from \texttt{test} split. The numbers of problems in each 
 of the 7 subjects are listed in \cref{tab:stat}. \cref{fig:hard_perturb} shows one example of our benchmark.

\begin{figure}[htbp]
        \centering
        \includegraphics[width=0.48\linewidth]{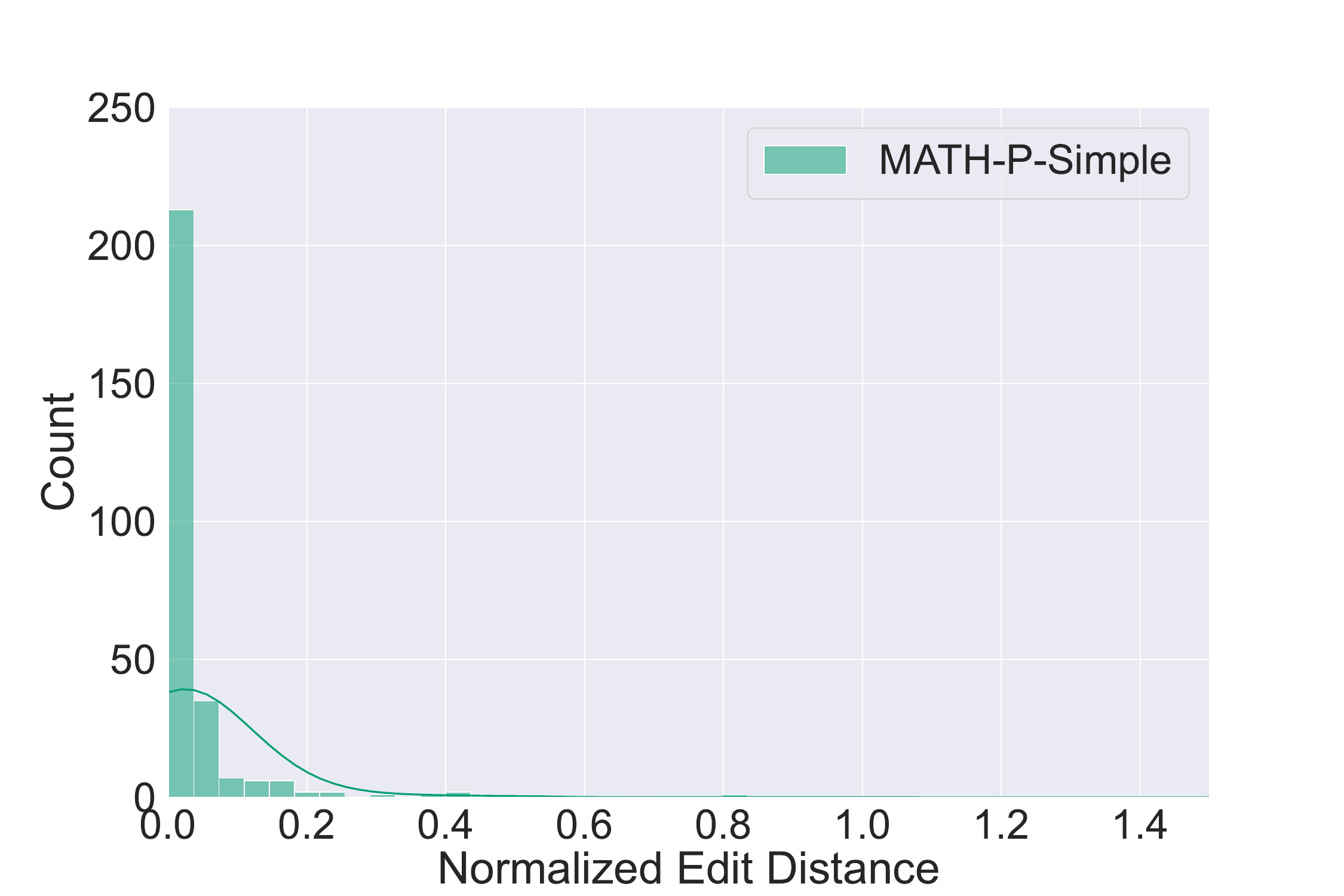}
        \includegraphics[width=0.48\linewidth]{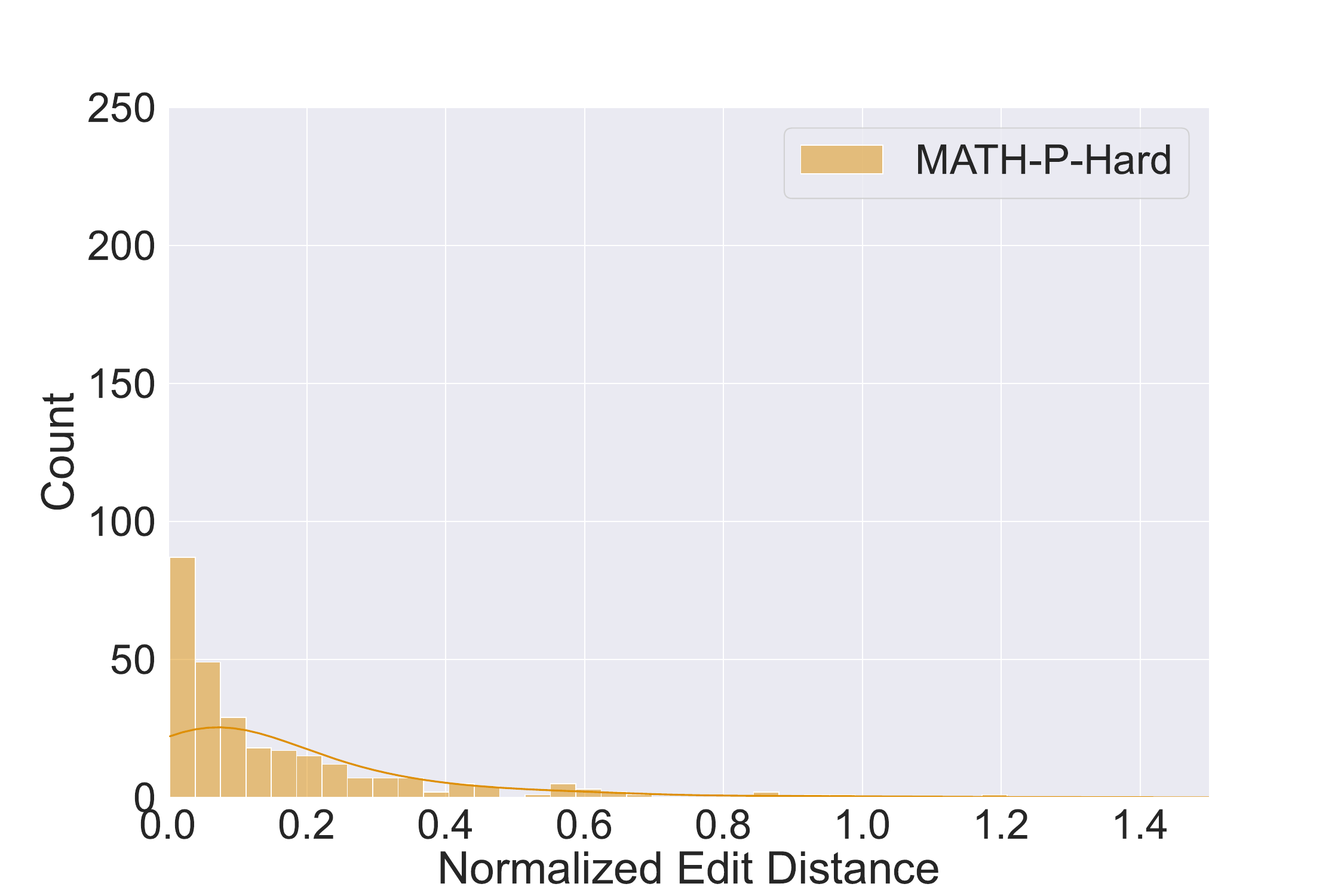}
        \includegraphics[width=0.48\linewidth]{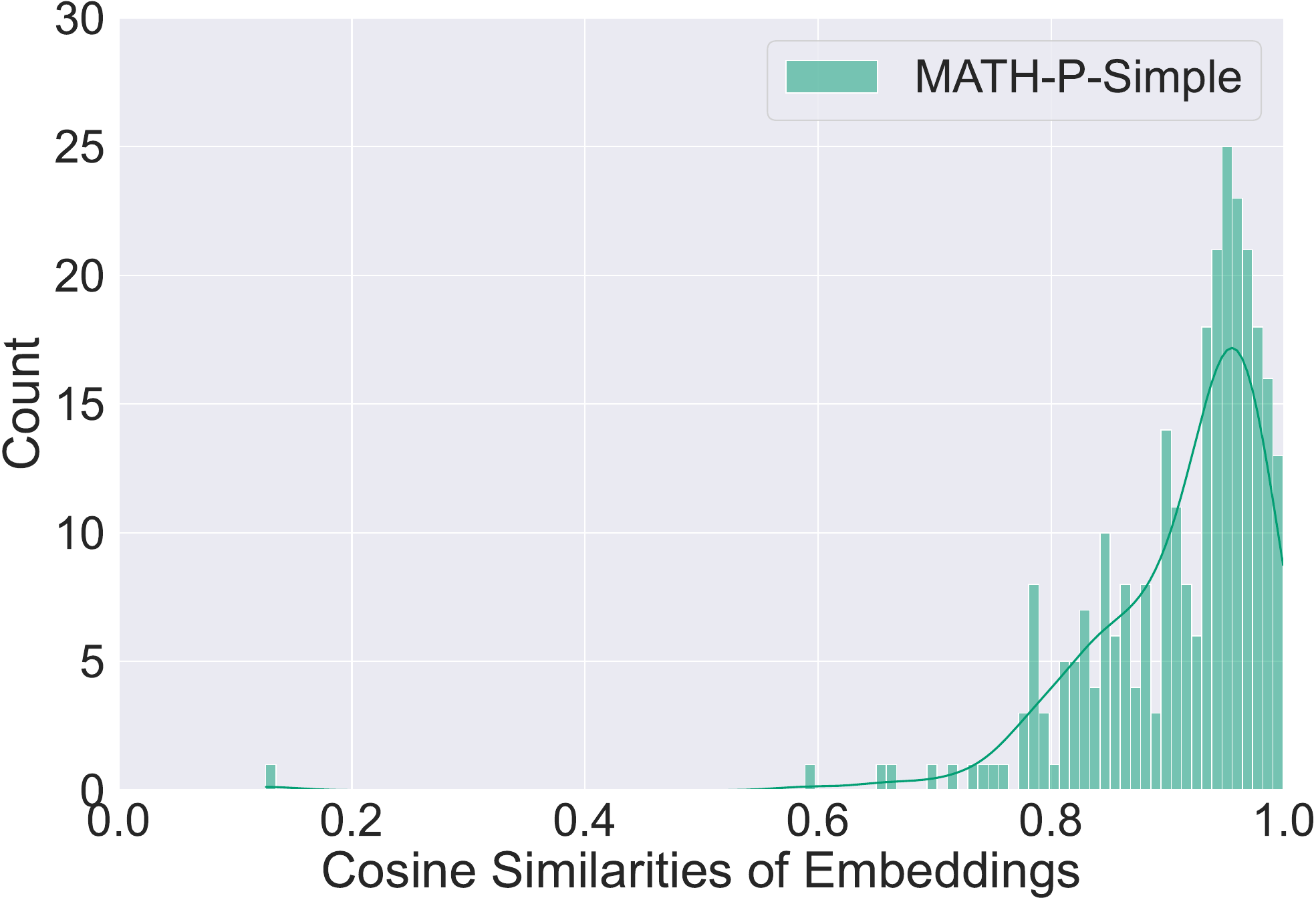}
        \includegraphics[width=0.48\linewidth]{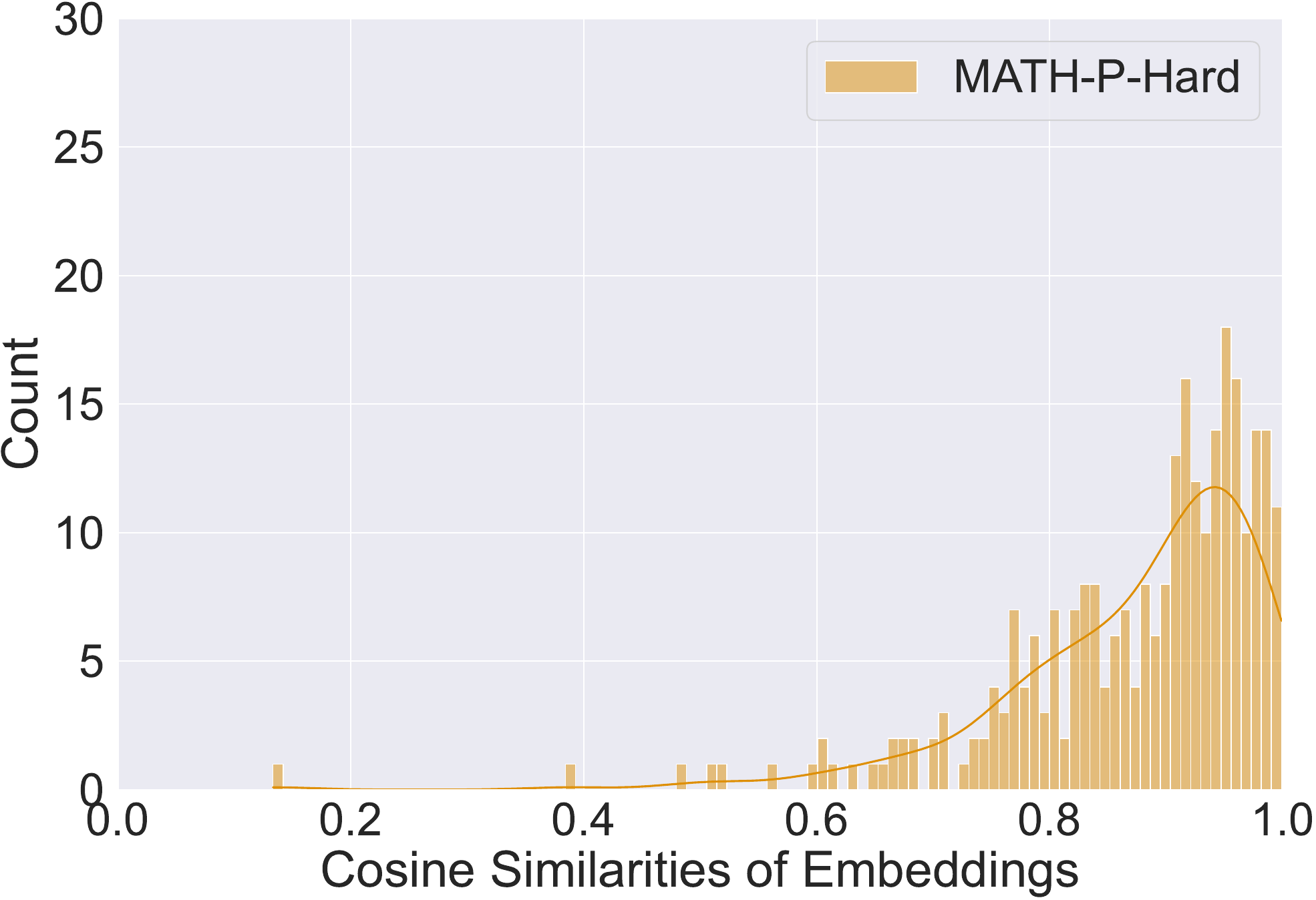}
        \caption{The distributions of edit distances and cosine similarities of embeddings between the perturbed problems and the original problems. The edit distances are normalized by the lengths of the original problems. The embedding model is OpenAI's \texttt{text-embedding-3-large}. 
        }
        \label{fig:edit_distance_cos_sim}
    \end{figure}

To quantify how similar the original problem and the modified problem are,  first, we calculate the edit distance between the modified problem and the original problem, normalized by the length of the original problem. Besides, we compute the cosine similarities between the embeddings of the two problems, where we use OpenAI's \texttt{text-embedding-3-large} embedding model. The distributions of the normalized edit distance and the cosine similarities are shown in \cref{fig:edit_distance_cos_sim}.

We also calculate the Mean Reciprocal Ranks (MRRs) when using the perturbed problem as the query to retrieve the corresponding original problem from the set of all 279 original problems, with the cosine similarities of embeddings being the ranking method. The MRRs of the \SAME problems and \HARD problems are 0.995 and 0.986, respectively, indicating that the corresponding original problem and solution are likely to be retrieved using typical semantic-based retrieval methods.

\textbf{Common Strategies for Perturbations.}

For \SAME, most of the problems are modified numerically without making the problems fundamentally different. Our annotators have checked these numerical modifications are non-essential to the problems, so the modified problems can be solved with the same reasoning patterns. 
Besides, our annotators also adopt other types of changes. For example, asking for a different but related quantity, adding/removing non-essential constraints, and changing a mathematical concept to its contrasting counterpart.

For \HARD, the modification strategies are more diverse and problem-specific. A general strategy is to increase the complexity of the mathematical objects involved. For example, raising the degrees of polynomials will make them harder to solve or factorize. Altering key numbers to large values can make brute-force solutions infeasible. Instead, solving the problem requires deriving general formulas or applying deeper theorems rather than relying on computational shortcuts. Other common strategies include relaxing constraints to cover more general cases, changing essential conditions so the original simplifying properties (e.g. symmetry, reducibility, linearity) no longer hold.

\section{Experimental Results}

    \begin{table*}[htbp]
    
    \caption{Zero-shot CoT performance of the LLMs (accuracy, \%). \Original refers to the set of 279 unmodified problems.  For the \texttt{train} and \texttt{test} columns, we report the accuracies for problems that \textit{originate} from the \texttt{train} split and \texttt{test} split, respectively. }
    \centering
    \resizebox{0.8\textwidth}{!}{
    \begin{tabular}{l>{\columncolor{gray!10}}ccc>{\columncolor{gray!10}}ccc>{\columncolor{gray!10}}ccc}
     \toprule
    \multirow{2}{*}{\textbf{Model}}   & \multicolumn{3}{c}{\textbf{\Original}}  & \multicolumn{3}{c}{\textbf{\SAME}} & \multicolumn{3}{c}{\textbf{\HARD}}   \\ 
    \cmidrule(r){2-4} \cmidrule(r){5-7} \cmidrule(r){8-10}
    &   All & \texttt{train} & \texttt{test} &  All & \texttt{train} & \texttt{test} & All & \texttt{train} & \texttt{test} \\
    \midrule
    Gemini-2.0-flash-thinking-exp &   92.47 & 92.68 & 92.17  &   91.04 & 87.80 & 95.65  &   78.14 & 77.44 & 79.13  \\ 
    o1-preview &   87.81 & 88.41 & 86.96  &   87.81 & 87.80 & 87.83  &   72.40 & 73.78 & 70.43  \\ 
    o1-mini &   94.27 & 93.90 & 94.78  &   94.98 & 93.29 & 97.39  &   78.49 & 79.27 & 77.39  \\ 
    \midrule
    Gemini-2.0-flash-exp &   88.17 & 87.20 & 89.57  &   82.80 & 81.71 & 84.35  &   67.03 & 68.29 & 65.22  \\ 
    Gemini-1.5-pro &   77.78 & 77.44 & 78.26  &   77.42 & 76.83 & 78.26  &   56.63 & 56.10 & 57.39  \\ 
    GPT-4o &   67.03 & 68.90 & 64.35  &   62.01 & 60.98 & 63.48  &   39.43 & 37.80 & 41.74  \\ 
    GPT-4-turbo &   56.99 & 55.49 & 59.13  &   55.20 & 56.71 & 53.04  &   34.41 & 36.59 & 31.30  \\ 
    Claude-3.5-Sonnet &   64.52 & 62.80 & 66.96  &   58.42 & 57.32 & 60.00  &   38.71 & 38.41 & 39.13  \\ 
    Claude-3-Opus &   41.94 & 39.02 & 46.09  &   41.94 & 39.63 & 45.22  &   26.52 & 25.00 & 28.70  \\ 
    \midrule
    Llama-3.1-8B-Instruct &   36.56 & 45.12 & 24.35  &   31.54 & 35.37 & 26.09  &   10.04 & 10.98 & 8.70  \\ 
    Gemma-2-9b-it &   27.60 & 28.05 & 26.96  &   27.60 & 30.49 & 23.48  &   11.83 & 12.80 & 10.43  \\ 
    Phi-3.5-mini-instruct &   26.16 & 27.44 & 24.35  &   28.67 & 26.83 & 31.30  &   14.34 & 15.24 & 13.04  \\ 
    \midrule
    Deepseek-math-7b-rl &   37.28 & 42.68 & 29.57  &   33.33 & 35.37 & 30.43  &   13.62 & 15.85 & 10.43  \\ 
    Qwen2.5-Math-7B-Instruct &   58.78 & 59.15 & 58.26  &   51.61 & 50.00 & 53.91  &   27.24 & 29.88 & 23.48  \\ 
    Mathstral-7b-v0.1 &   36.56 & 43.29 & 26.96  &   36.20 & 42.07 & 27.83  &   14.70 & 16.46 & 12.17  \\ 
    NuminaMath-7B-CoT &   43.73 & 51.22 & 33.04  &   40.14 & 44.51 & 33.91  &   17.20 & 18.90 & 14.78  \\ 
    MetaMath-13B-V1.0 &   21.15 & 32.32 & 5.22  &   7.53 & 7.32 & 7.83  &   5.73 & 4.88 & 6.96  \\ 
    MAmmoTH2-8B &   12.90 & 11.59 & 14.78  &   17.92 & 17.07 & 19.13  &   7.53 & 10.37 & 3.48  \\ 
    \bottomrule
    \end{tabular}
    }
    
    \label{tab:main}
    \end{table*}

\textbf{Evaluation Setting.} We adopt zero-shot chain-of-thought (CoT)~\citep{wei2022chain, kojima2022large} as the standard evaluation method on our benchmarks. For comparison, we also evaluate the  models on the set of the original 279 problems, referred to as \Original in the following subsections. We do not allow any tool usage including access to a code interpreter, as we find that many problems can be trivially solved by writing a brute-force search program. 

To check whether the generated answer matches the ground-truth answer, we adopt an equivalence checker following~\citet{hendrycksmath2021, shao2024deepseekmath}, which first performs string normalization and then uses sympy package to check the equivalence of two mathematical objects.

\subsection{Benchmarking the performance of LLMs}
\label{sec:benchmarking}

We consider a wide range of language models including long-CoT models, closed-sourced large models, open-sourced small models, and math-specific models. The version information of the models is deferred to Appendix~\ref{appendix:model}.

In \cref{tab:main}, we report the overall accuracies of the LLMs on Original, \SAME, and \HARD, and also separately calculate the accuracies for problems that originate from the \texttt{train} split and \texttt{test} split.
As expected, for all the models we evaluate, we find that the performance on \HARD is significantly lower than the original problems, which indicates \HARD is more difficult. 

In the meantime, most models also suffer a slight performance drop on \SAME compared to the original problems. We note that the performance drops mainly come from the \texttt{train} split. Generalization errors still exist for the state-of-the-art models even when the test examples follow the exact same reasoning patterns as the training problems.

For problems that originate from the \texttt{test} split, ideally, both the original problem and its \SAME modification should be equally ``unseen'' to the model. We observe mixed results empirically from \cref{tab:main}: for gemini-2.0-flash-exp, GPT-4-turbo, claude-3.5-sonnet, the performance drops are larger than 5\%, while surprisingly the performance of Phi-3.5-mini-instruct increases. For most of the models we evaluated, the accuracies on \SAME \texttt{test} split are close to the accuracies on the original \texttt{test} split. 
We commend that while \citet{srivastava2024functional} found a \textit{relatively} 58\% to 80\% performance drop between their modified benchmark and the original MATH benchmark among a different set of the models (the best model they tested was GPT-4), we did not observe such huge gaps for the models we evaluate, which is a sign of the progress in the robustness of the newly developed models against simple perturbations.

\textbf{Inference-time Scaling.}  Scaling inference-time computes has been shown to be able to boost the performance of LLMs~\citep{wang2022self, brown2024large, wu2024empirical, cobbe2021training, lightman2023let}. We defer the study of inference-time scaling on our benchmarks to \cref{sec:inference:scaling}.

\subsection{Failure Mode Analysis}
\label{sec:failure:mode}

\begin{figure*}[t]
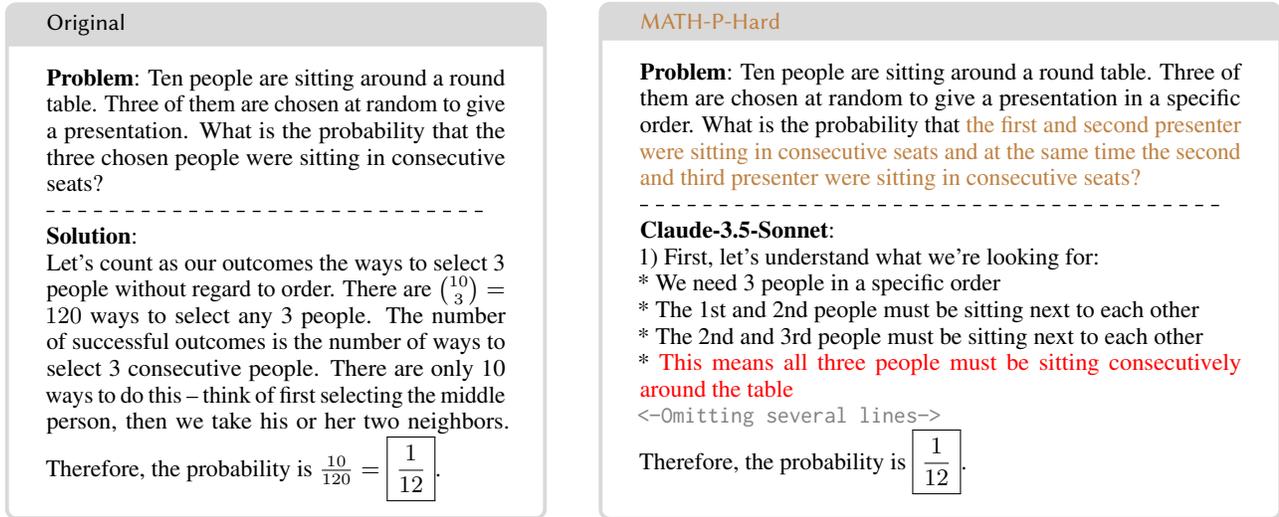

\begin{small}
    \centering
    \begin{minipage}[c]{0.42\textwidth}
        \begin{tcolorbox}[colback=white, colframe=gray!30, title=\textcolor{black}{\textsf{Original}}]
        \textbf{Problem}: Ten people are sitting around a round table.  Three of them are chosen at random to give a presentation.  What is the probability that the three chosen people were sitting in consecutive seats?

        \dashline

        \textbf{Solution}:

        Let's count as our outcomes the ways to select 3 people without regard to order.  There are $\binom{10}{3} = 120$ ways to select any 3 people.  The number of successful outcomes is the number of ways to select 3 consecutive people.  There are only 10 ways to do this -- think of first selecting the middle person, then we take his or her two neighbors.  Therefore, the probability is $\frac{10}{120} = \boxed{\frac{1}{12}}$.
        \vspace{-1mm}
        \end{tcolorbox}

    \end{minipage}
    \hfill
    \begin{minipage}[c]{0.53\textwidth}
        \begin{tcolorbox}[colback=white, colframe=gray!30, title=\HARD]
            \textbf{Problem}: Ten people are sitting around a round table.  Three of them are chosen at random to give a presentation in a specific order.  What is the probability that \textcolor{brown}{the first and second presenter were sitting in consecutive seats and at the same time the second and third presenter were sitting in consecutive seats?}

            \dashline

            \textbf{Claude-3.5-Sonnet}: \\
            1) First, let's understand what we're looking for: \\
           * We need 3 people in a specific order \\
           * The 1st and 2nd people must be sitting next to each other \\
           * The 2nd and 3rd people must be sitting next to each other \\
           * \textcolor{red}{This means all three people must be sitting consecutively around the table} \\
    \textcolor{gray}{\texttt{<--Omitting several lines-->}}

Therefore, the probability is $\boxed{\frac{1}{12}}$. %
        \end{tcolorbox}
        
    \end{minipage}
\end{small}
    \caption{An example of memorization \textit{coupled with} incorrect reasoning: The model incorrectly reduces the modified condition to the original condition, and then follows the original reasoning pattern. The correct answer is $\boxed{\nicefrac{1}{36}}$. {We manually performed 20 repeated trials and found that Claude-3.5-Sonnet has a pass rate of 50\%. Among the mistakes, 30\% are due to the memorization issue above.}}
    \label{fig:case:2}
    \vspace{-3mm}
\end{figure*}

\begin{figure*}[ht]
\begin{small}
    \centering
    \begin{minipage}[c]{0.54\textwidth}
        \begin{tcolorbox}[colback=white, colframe=gray!30, title=\textcolor{black}{\textsf{Original}}]
        \textbf{Problem}: If $\left\lfloor n^2/4 \right\rfloor - \lfloor n/2 \rfloor^2 = 2$, then find all integer values of $n$.
        
        \dashline
        
        \textbf{Solution:}
        
        If $n$ is even, then we can write $n = 2m$ for some integer $m$. Substituting, $$\left \lfloor (2m)^2/4 \right\rfloor - \left\lfloor (2m)/2 \right\rfloor^2 = m^2 - m^2 = 0.$$Hence, $n$ must be odd; we can write $n = 2m+1$ for some integer $m$. Substituting,   
        \begin{align*}
        &\left \lfloor (2m+1)^2/4 \right. \rfloor - \left\lfloor (2m+1)/2 \right\rfloor^2\\
        & \text{\textcolor{gray}{\texttt{<--Omitting several lines-->}}}\\
        & = m.
        \end{align*}Thus, we find $m = 2$ and $n = \boxed{5}$ as the unique integer solution.
        \vspace{1mm}
        \end{tcolorbox}
    \end{minipage}
    \hfill
    \begin{minipage}[c]{0.44\textwidth}
        \begin{tcolorbox}[colback=white, colframe=gray!30, title=\HARD]
            \textbf{Problem}: If \textcolor{brown}{$\left\lfloor n^2/9 \right\rfloor - \lfloor n/3 \rfloor^2 = 2$}, then find \textcolor{brown}{the smallest} integer value of $n$.

        \dashline

        \textbf{o1-mini:} 
        
        Let's solve the equation step-by-step:
        
        \textcolor{gray}{\texttt{<--Omitting several lines-->}}

5. **Analyze for each value of \(r\):**

    \textcolor{gray}{\texttt{<--Omitting several lines-->}}
    
   - **For \(r = 1\):**
     \[
     2 \leq \frac{2k}{3} + \frac{1}{9} < 3
     \]
     
     Solving gives \(k = 3\) and \(k = 4\). Therefore, \(n = 10\) and \(n = 13\).

\textcolor{gray}{\texttt{<--Omitting several lines-->}}

**Final Answer:**

\[\boxed{\textcolor{red}{10 \text{ and } 13}}\] 
        \end{tcolorbox}
    \end{minipage}
\end{small}
    \caption{An example of memorizing the desired outcome. The model outputs all integer values instead of the smallest integer value. The correct answer is $\boxed{10}$. {We manually performed 20 repeated trials and found that o1-mini has a pass rate of 75\%. All the 25\% errors are due to this specific memorization issue above.}}
    \label{fig:case:1}
    \vspace{-3mm}
\end{figure*}

To study \after{the generalization abilities of models against hard perturbations} , we focus on the set of problems where the models fail on the \HARD modification but correctly solve either the original problem or the \SAME modification, which accounts for 20\%-47\% of the total problems. For these problems, one can use the correct solutions to the easier problems as a reference to better determine the failure modes on the hard problems. We defer the discussion on the other cases to Appendix~\ref{appendix:category}.

First, we observe general failure modes when models are exposed to harder problems, including making mistakes in basic numerical computations and algebraic operations, making unjustified claims, missing several cases, and lacking certain math knowledge. These types of errors are more prominent in weaker models.

Besides general failure modes, when we compare the wrong solution to the \HARD modification with the solutions to the easier versions, we are able to recognize an adequate number of \red{memorization issues}. Specifically, we found that models may \textbf{ignore the modified assumptions and presume that the original assumptions still hold}; see \cref{fig:case:2} for an example. In other cases, the models may \textbf{blindly apply the techniques for the original problems} without first determining whether these techniques are still suitable in the modified setting (the responses in \cref{fig:hard_perturb} are such an example generated by GPT-4o). Interestingly, the models may even \textbf{output the desired outcome of the original problem} (not provided in the context) instead of the modified problem, e.g. \cref{fig:case:1}. This kind of memorization behavior is difficult to capture with most existing type of perturbations in the literature (similar to our \SAME) that does not require different solving strategies.

    These issues are often coupled with other types of errors and pervasive among the models we evaluated. 
    For large models, we estimate the percentages of errors caused by memorization to be 40\% for o1-mini and 25\% 
    \begin{wrapfigure}{r}{0.12\textwidth}
        \centering
        \includegraphics[width=\linewidth]{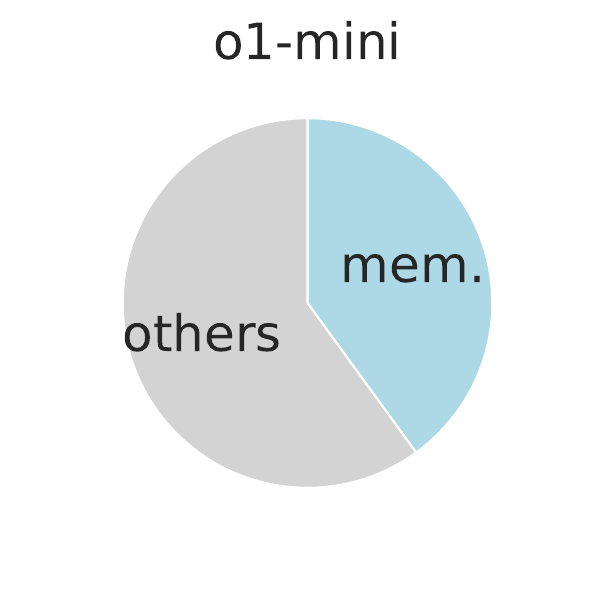}
        \vspace{4mm}
        \includegraphics[width=\linewidth]{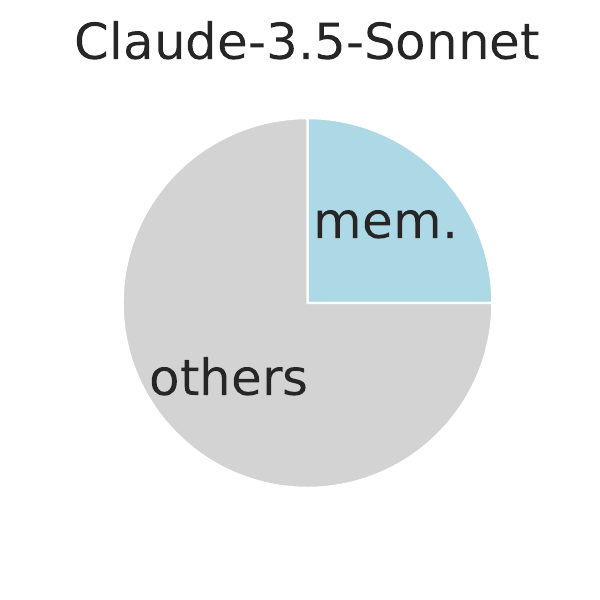} 
        \vspace{-2mm}
        \label{fig:memo_percentage}
    \end{wrapfigure}
    for Claude-3.5-Sonnet, via manual inspections of 20 error cases.
    The general failure modes due to insufficient capabilities are largely reduced for stronger models, making the \red{memorization issues} more prominent. 
    As the capabilities of language models continue to advance, we expect the memorization issues will be the next bottleneck of reasoning models, and we urge more studies on investigating \after{the generalization abilities of reasoning models against hard perturbations.}

\subsection{Is Mode Collapse a Problem?}
\label{sec:naive:memorization}

We investigate whether the model makes errors due to \red{mode collapse}, which means the model fails to identify the difference between the perturbed problem and the original problem (seen during its training time) and the model's response \textit{collapses} to the response to the original problem with the identical answer.

For each model, we report $n_{\text{same}}$, the number of problems where the model's final answer coincides with the ground-truth answer of the corresponding original problem. For those responses, we also compute the edit distance between the full response to the modified problem and the full response to the original problem. The full result is deferred to \cref{tab:naive_memorization} in the appendix.

We see that this type of failure mode accounts for less than 10\% of the total errors except for three models (gemini-2.0-flash-thinking-exp, o1-mini, and gemini-2.0-flash-exp) on \HARD. After manual inspection, we find that except for only 1 problem pair where gemma-2-9b-it generates the identical answer for the original problem and the modified problem, we do not see collapses of the outputs \textbf{in the superficial text form}. Therefore, we conclude that naive recitation of the \textit{training material} is not the major reason for producing the same answers. \textit{Instead}, the model's responses to the modified problems often collapse to the responses to the original problems \textbf{in more subtle manners}, e.g. ignoring or failing to understand the modified assumptions; see \cref{fig:case:2} for an example.

\subsection{Does In-context Learning Help or Hurt?}
\label{sec:icl}

\begin{figure*}[htbp]
    \centering
    \includegraphics[width=0.33\linewidth]{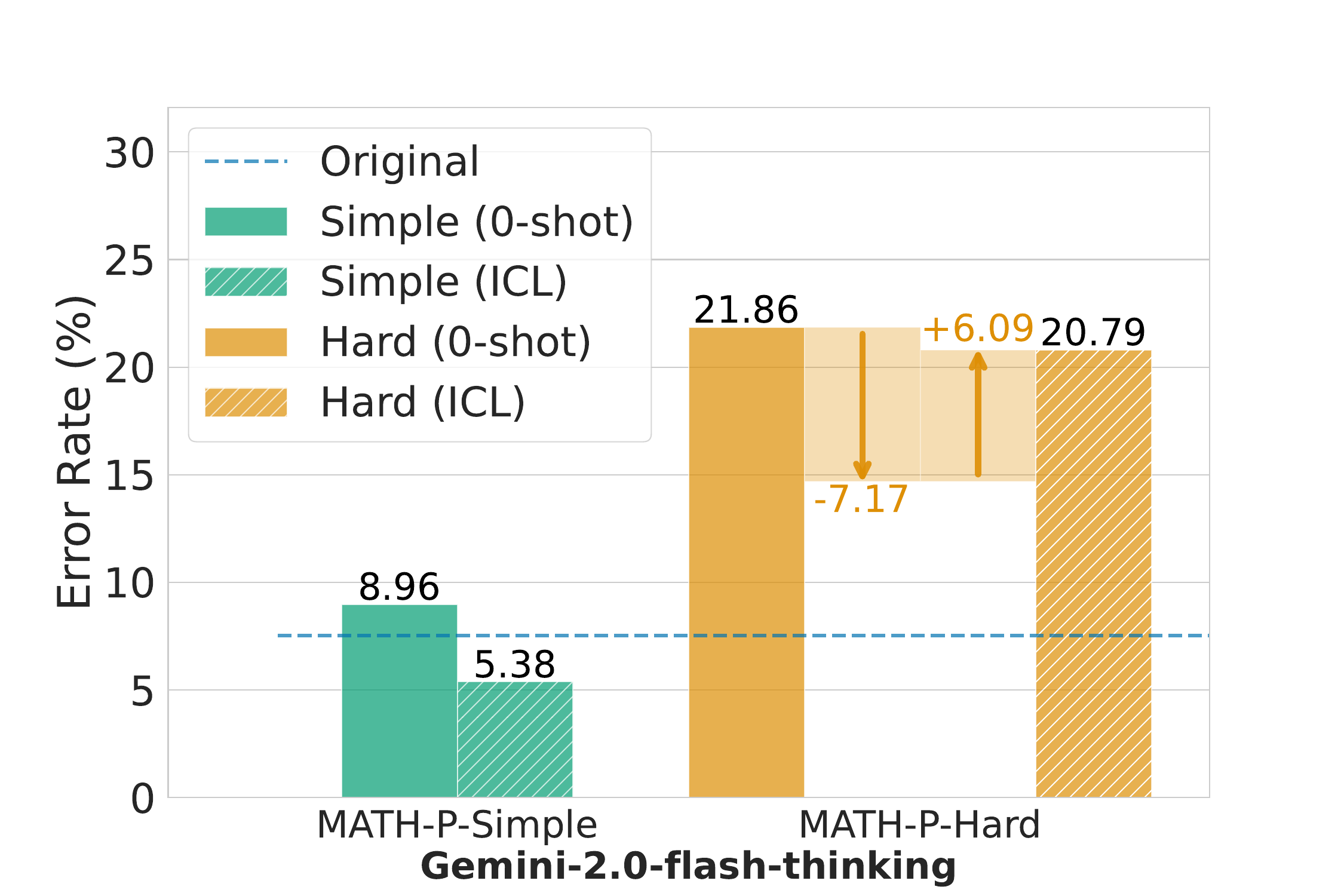}
    \includegraphics[width=0.32\linewidth]{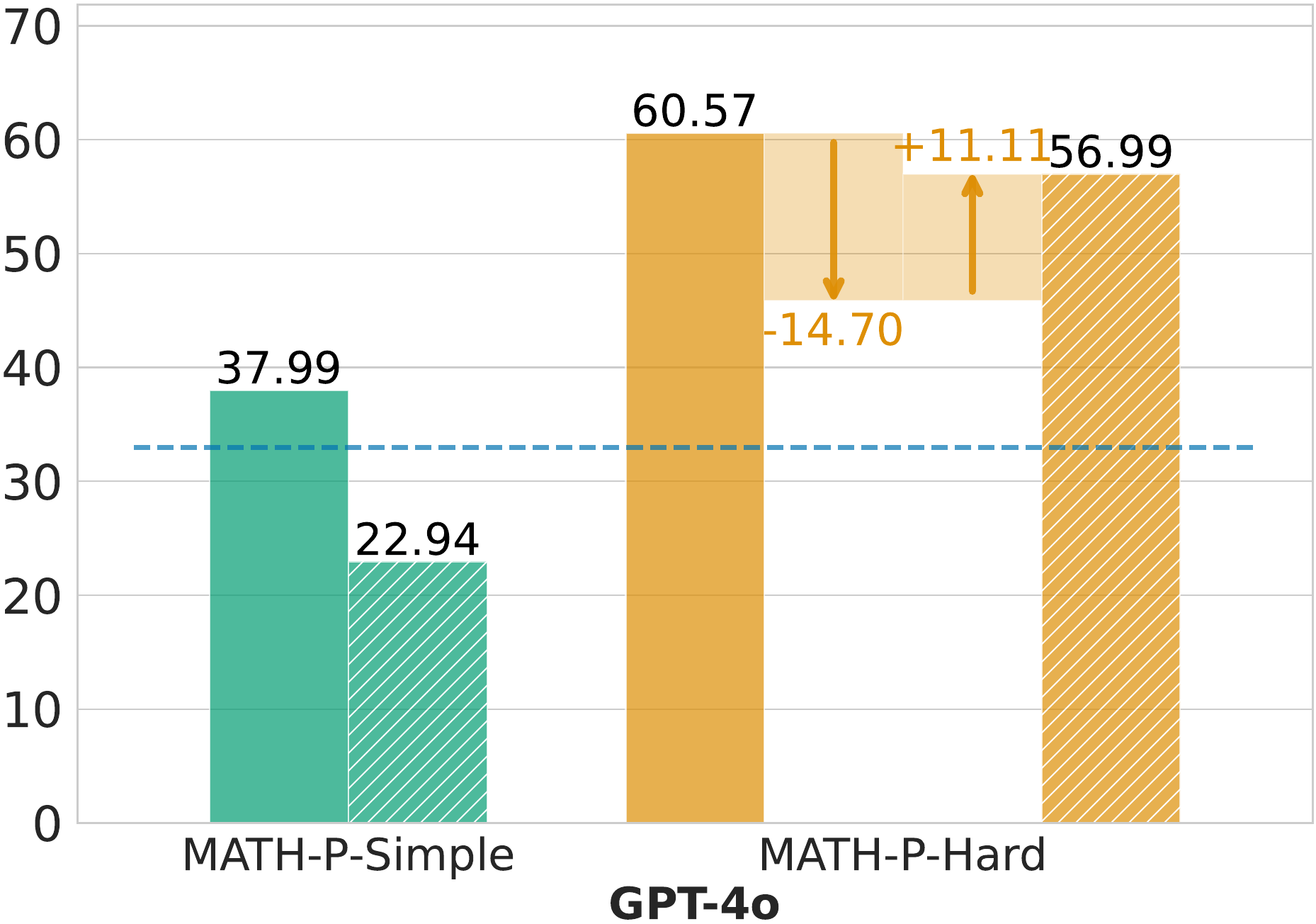}
    \includegraphics[width=0.325\linewidth]{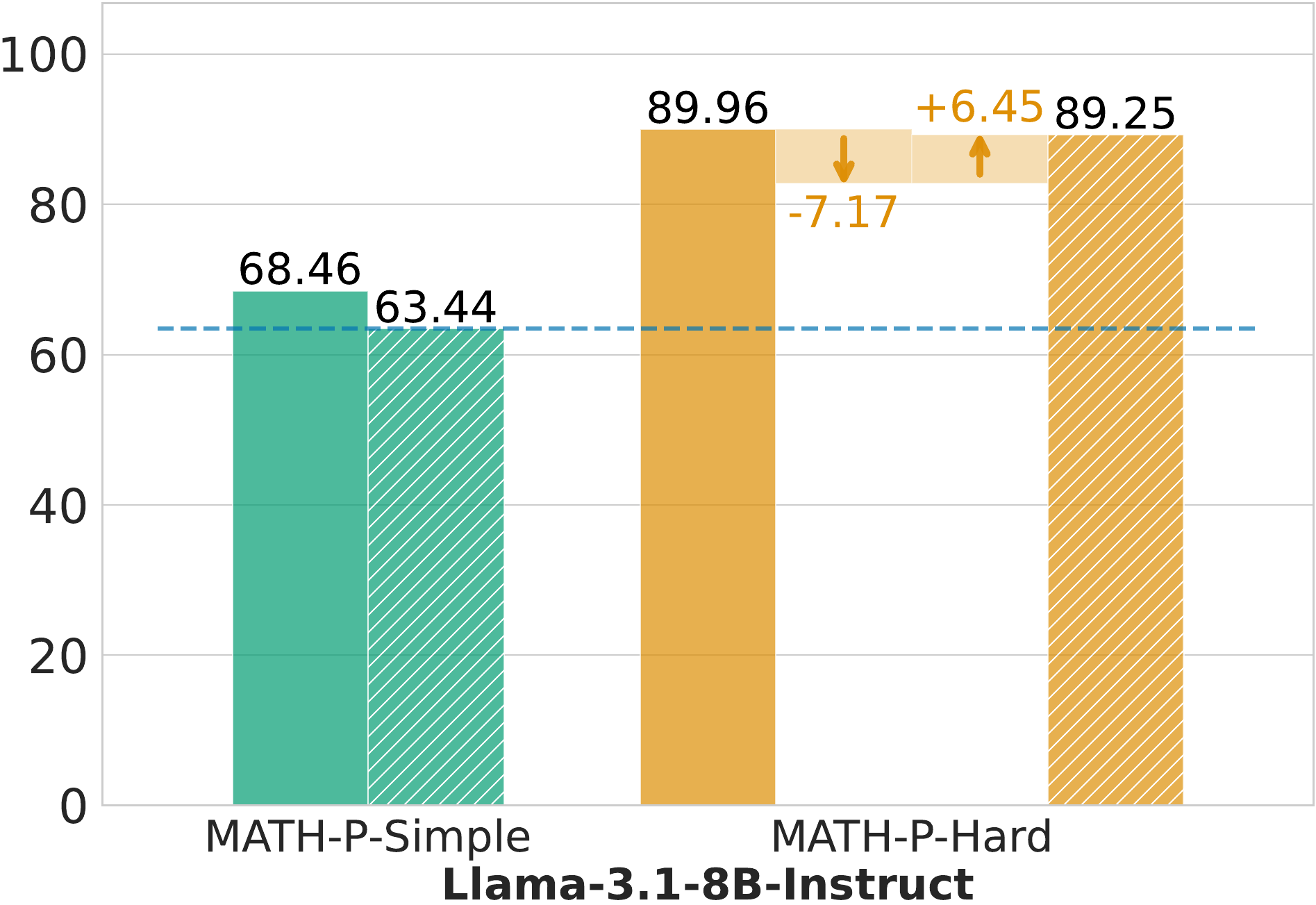}
    \caption{The error rates (\%) of the models without and with the original problem and solution as the in-context learning (ICL) example. For \HARD, we decompose the influences of in-context learning into \textbf{ICL effect} (the down arrow $\textcolor{brown}{\boldsymbol{\downarrow}}$), which reduces the error rates, and \textbf{misleading effect} (the up arrow $\textcolor{brown}{\boldsymbol{\uparrow}}$), which increases the error rates. %
    }
    \label{fig:icl}
    \vspace{-2mm}
\end{figure*}

In this subsection, we investigate whether using the corresponding original unmodified problem and solution as the one-shot in-context learning (ICL) example will help with the modified problems in \SAME and \HARD. We visualize the influences of ICL for three models in \cref{fig:icl} and defer the full result to \cref{tab:OIC}.

As expected, using the original (problem, solution) pair as a one-shot in-context demonstration boosts the performance of nearly all the models on \SAME, which should be solvable by simply applying the original solution steps to the modified setting.   

As for the \HARD modifications, there are two factors that need to be considered: (1) \textbf{ICL effect}: the original solutions may supply the model with desired mathematical knowledge that is also helpful for solving the modified problems; (2) \textbf{misleading effect}: on the other hand, as there are subtle differences between the original problems and the \HARD modifications, the models may fail to recognize such differences and be misled by the demonstrated solutions.
Accordingly, in \cref{tab:icl:breakdown} and \cref{fig:icl}, and we calculate and visualize (1) $n_{\text{wrong} \to \text{correct}}$, the number of problems that initially the model fails on \textit{without} the in-context demonstrations but answers correctly \textit{with} the in-context demonstrations, and (2) $n_{\text{correct} \to \text{wrong}}$, the number of problems that initially the model answers correctly \textit{without} demonstrations but fails on \textit{with} demonstrations. 

We observe that many \HARD problems become solvable with the original problems and solutions as demonstrations. The percentages to the number of total errors without demonstrations are larger for closed-sourced large models (24\%-40\%) and smaller for open-sourced small models (2\%-15\%), due to their differences in mathematical capabilities and in-context learning capabilities. \textbf{However}, we also observe many \HARD problems become incorrect with demonstrations, and the percentages are higher for large models (18\%-40\%) than small models (4\%-15\%). 
The misleading effect counteracts the effect of in-context learning, leaving only marginal improvements (less than 5\%) on the \HARD for most models.

As in-context learning can be viewed as a form of (test-time) training, we hypothesize that any naive fine-tuning technique with a limited distribution of problem settings will hurt the \after{generalization} of the language models against hard perturbations. %

\section{Related Work}

\textbf{Perturbations to Existing Mathematical Benchmarks.} There is a considerable amount of work focusing on performing perturbations to existing mathematical benchmarks.  
\citet{shi2023large} built GSM-IC from GSM8K~\citep{cobbe2021training} by adding irrelevant context to the problem. GSM-Plus~\citep{li2024gsm} creates 8 types of variations to each of the GSM8K problem and ensure that the perturbed problem is of the same difficulty.
\citet{mirzadeh2024gsm} built GSM-Symbolic that alters the numerical values and entity names via symbolic templates of both the problems and the solution steps. 
Similarly, Functional MATH~\citep{srivastava2024functional} is created from the MATH dataset~\citep{hendrycksmath2021}, and Putnam-AXIOM~\citep{gulati2024putnamaxiom} from the Putnam Mathematical Competition.

This line of work performed \textbf{simple perturbations} to existing mathematical benchmarks and the perturbed problems can be solved with the same solution steps and the same reasoning pattern as the original ones. In contrast, we performed \textbf{hard perturbations} to curate \HARD, where the original reasoning pattern does not apply.

\textbf{Memorization.} Memorization is a well-studied phenomenon in machine learning~\citep{feldman2020neural, zhang2021understanding, feldman2020does} and has become increasingly prevalent in large language models, due to the growing of the pretraining corpora and the scaling of the model sizes. 
Verbatim memorization, i.e., recitation of the training material, has significant potential consequences ranging from privacy violations~\citep{carlini2022privacy, brown2022does,huang2023privacy} and copyright infringement~\citep{shi2023detecting, karamolegkou2023copyright, wei2024evaluating, chen2024copybench} to training data security risks~\citep{carlini2021extracting, nasr2023scalable}.
Prior work has investigated various factors influencing verbatim memorization, including sequence duplicates~\citep{lee2021deduplicating, hernandez2022scaling}, model size~\citep{tirumala2022memorization}, and sequence position~\citep{biderman2023pythia}.

In contrast, we investigate the effect of memorization within the mathematical reasoning context. Our methodology falls into the category of \textit{counterfactual tests}~\citep{zhang2023counterfactual, wu2023reasoning, zheng2023large,  xie2024memorization}, where we construct perturbed problems different from the existing ones to test the \after{generalization} of LLMs and examine memorization effects.
Through extensive case studies, we find that LLMs can exhibit subtle forms of memorization \textit{beyond} naive verbatim memorization.

\textbf{Comparison with MATH$^2$~\citep{shah2024ai}.}
\citet{shah2024ai} created MATH$^2$ by combining random pairs of skills extracted from MATH~\citep{hendrycksmath2021} to generate harder problems that require both skills to solve. Their benchmark is mathematically harder, but there are no natural ``original problems'' as references. Therefore, MATH$^2$ is not directly suitable for investigating the memorization effects of language models. In comparison, our \HARD are modified directly from the problems in MATH so that the modified problems require harder skills to solve. \HARD can serve as both a harder math benchmark and a testbed to investigate memorizations of LLMs.

\section{Conclusion}
In this work, we study the \after{generalization} of large language models' math reasoning abilities against hard perturbations of the problems. We modified 279 problems from the level-5 problems of the MATH dataset~\citep{hendrycksmath2021} into \SAME (used for control experiments) and \HARD, via simple perturbations and hard perturbations, respectively. We found performance degradations of all models on \HARD, and many of the errors can be traced to a new form of memorization, where the model memorizes the problem-solving techniques from the training set and blindly applies them without judging whether the modified settings are still suitable. Using the original unmodified problem and solution for in-context learning can deteriorate this issue. We expect the \after{generalization} against hard perturbations to be the next major bottleneck of LLMs' reasoning abilities and urge future work in this direction.

\section*{Acknowledgements}
We acknowledge Professor Jonathan Cohen (Princeton) and Andrew Tomkins (Google) for the helpful feedback and discussion.
Kaixuan Huang acknowledges the support of Google PhD Fellowship.
Chi Jin acknowledges the support from the National Science Foundation NSF-OAC-2411299 and NSF-IIS-2239297.
Mengdi Wang acknowledges support by NSF grants DMS-1953686, IIS-2107304, and ONR grant 1006977.
The research is also supported by Princeton Language and Intelligence (PLI) Compute Cluster.

\bibliography{main}

\begin{thebibliography}{62}
\providecommand{\natexlab}[1]{#1}
\providecommand{\url}[1]{\texttt{#1}}
\expandafter\ifx\csname urlstyle\endcsname\relax
  \providecommand{\doi}[1]{doi: #1}\else
  \providecommand{\doi}{doi: \begingroup \urlstyle{rm}\Url}\fi

\bibitem[Abdin et~al.(2024)Abdin, Aneja, Awadalla, Awadallah, Awan, Bach, Bahree, Bakhtiari, Bao, Behl, Benhaim, Bilenko, Bjorck, Bubeck, Cai, Cai, Chaudhary, Chen, Chen, Chen, Chen, Chen, Cheng, Chopra, Dai, Dixon, Eldan, Fragoso, Gao, Gao, Gao, Garg, Giorno, Goswami, Gunasekar, Haider, Hao, Hewett, Hu, Huynh, Iter, Jacobs, Javaheripi, Jin, Karampatziakis, Kauffmann, Khademi, Kim, Kim, Kurilenko, Lee, Lee, Li, Li, Liang, Liden, Lin, Lin, Liu, Liu, Liu, Liu, Liu, Luo, Madan, Mahmoudzadeh, Majercak, Mazzola, Mendes, Mitra, Modi, Nguyen, Norick, Patra, Perez-Becker, Portet, Pryzant, Qin, Radmilac, Ren, de~Rosa, Rosset, Roy, Ruwase, Saarikivi, Saied, Salim, Santacroce, Shah, Shang, Sharma, Shen, Shukla, Song, Tanaka, Tupini, Vaddamanu, Wang, Wang, Wang, Wang, Wang, Wang, Ward, Wen, Witte, Wu, Wu, Wyatt, Xiao, Xu, Xu, Xu, Xue, Yadav, Yang, Yang, Yang, Yang, Yu, Yuan, Zhang, Zhang, Zhang, Zhang, Zhang, Zhang, Zhang, and Zhou]{abdin2024phi3technicalreporthighly}
Marah Abdin, Jyoti Aneja, Hany Awadalla, Ahmed Awadallah, Ammar~Ahmad Awan, Nguyen Bach, Amit Bahree, Arash Bakhtiari, Jianmin Bao, Harkirat Behl, Alon Benhaim, Misha Bilenko, Johan Bjorck, Sébastien Bubeck, Martin Cai, Qin Cai, Vishrav Chaudhary, Dong Chen, Dongdong Chen, Weizhu Chen, Yen-Chun Chen, Yi-Ling Chen, Hao Cheng, Parul Chopra, Xiyang Dai, Matthew Dixon, Ronen Eldan, Victor Fragoso, Jianfeng Gao, Mei Gao, Min Gao, Amit Garg, Allie~Del Giorno, Abhishek Goswami, Suriya Gunasekar, Emman Haider, Junheng Hao, Russell~J. Hewett, Wenxiang Hu, Jamie Huynh, Dan Iter, Sam~Ade Jacobs, Mojan Javaheripi, Xin Jin, Nikos Karampatziakis, Piero Kauffmann, Mahoud Khademi, Dongwoo Kim, Young~Jin Kim, Lev Kurilenko, James~R. Lee, Yin~Tat Lee, Yuanzhi Li, Yunsheng Li, Chen Liang, Lars Liden, Xihui Lin, Zeqi Lin, Ce~Liu, Liyuan Liu, Mengchen Liu, Weishung Liu, Xiaodong Liu, Chong Luo, Piyush Madan, Ali Mahmoudzadeh, David Majercak, Matt Mazzola, Caio César~Teodoro Mendes, Arindam Mitra, Hardik Modi, Anh Nguyen,
  Brandon Norick, Barun Patra, Daniel Perez-Becker, Thomas Portet, Reid Pryzant, Heyang Qin, Marko Radmilac, Liliang Ren, Gustavo de~Rosa, Corby Rosset, Sambudha Roy, Olatunji Ruwase, Olli Saarikivi, Amin Saied, Adil Salim, Michael Santacroce, Shital Shah, Ning Shang, Hiteshi Sharma, Yelong Shen, Swadheen Shukla, Xia Song, Masahiro Tanaka, Andrea Tupini, Praneetha Vaddamanu, Chunyu Wang, Guanhua Wang, Lijuan Wang, Shuohang Wang, Xin Wang, Yu~Wang, Rachel Ward, Wen Wen, Philipp Witte, Haiping Wu, Xiaoxia Wu, Michael Wyatt, Bin Xiao, Can Xu, Jiahang Xu, Weijian Xu, Jilong Xue, Sonali Yadav, Fan Yang, Jianwei Yang, Yifan Yang, Ziyi Yang, Donghan Yu, Lu~Yuan, Chenruidong Zhang, Cyril Zhang, Jianwen Zhang, Li~Lyna Zhang, Yi~Zhang, Yue Zhang, Yunan Zhang, and Xiren Zhou.
\newblock Phi-3 technical report: A highly capable language model locally on your phone, 2024.
\newblock URL \url{https://arxiv.org/abs/2404.14219}.

\bibitem[Achiam et~al.(2023)Achiam, Adler, Agarwal, Ahmad, Akkaya, Aleman, Almeida, Altenschmidt, Altman, Anadkat, et~al.]{achiam2023gpt}
Josh Achiam, Steven Adler, Sandhini Agarwal, Lama Ahmad, Ilge Akkaya, Florencia~Leoni Aleman, Diogo Almeida, Janko Altenschmidt, Sam Altman, Shyamal Anadkat, et~al.
\newblock Gpt-4 technical report.
\newblock \emph{arXiv preprint arXiv:2303.08774}, 2023.

\bibitem[Anthropic(2024)]{claude35}
Anthropic.
\newblock Claude-3-5-sonnet.
\newblock 2024.
\newblock URL \url{https://www.anthropic.com/news/claude-3-5-sonnet}.

\bibitem[Biderman et~al.(2023)Biderman, Schoelkopf, Anthony, Bradley, O’Brien, Hallahan, Khan, Purohit, Prashanth, Raff, et~al.]{biderman2023pythia}
Stella Biderman, Hailey Schoelkopf, Quentin~Gregory Anthony, Herbie Bradley, Kyle O’Brien, Eric Hallahan, Mohammad~Aflah Khan, Shivanshu Purohit, USVSN~Sai Prashanth, Edward Raff, et~al.
\newblock Pythia: A suite for analyzing large language models across training and scaling.
\newblock In \emph{International Conference on Machine Learning}, pages 2397--2430. PMLR, 2023.

\bibitem[Brown et~al.(2024)Brown, Juravsky, Ehrlich, Clark, Le, R{\'e}, and Mirhoseini]{brown2024large}
Bradley Brown, Jordan Juravsky, Ryan Ehrlich, Ronald Clark, Quoc~V Le, Christopher R{\'e}, and Azalia Mirhoseini.
\newblock Large language monkeys: Scaling inference compute with repeated sampling.
\newblock \emph{arXiv preprint arXiv:2407.21787}, 2024.

\bibitem[Brown et~al.(2022)Brown, Lee, Mireshghallah, Shokri, and Tram{\`e}r]{brown2022does}
Hannah Brown, Katherine Lee, Fatemehsadat Mireshghallah, Reza Shokri, and Florian Tram{\`e}r.
\newblock What does it mean for a language model to preserve privacy?
\newblock In \emph{Proceedings of the 2022 ACM conference on fairness, accountability, and transparency}, pages 2280--2292, 2022.

\bibitem[Bubeck et~al.(2023)Bubeck, Chandrasekaran, Eldan, Gehrke, Horvitz, Kamar, Lee, Lee, Li, Lundberg, et~al.]{bubeck2023sparks}
S{\'e}bastien Bubeck, Varun Chandrasekaran, Ronen Eldan, Johannes Gehrke, Eric Horvitz, Ece Kamar, Peter Lee, Yin~Tat Lee, Yuanzhi Li, Scott Lundberg, et~al.
\newblock Sparks of artificial general intelligence: Early experiments with gpt-4.
\newblock \emph{arXiv preprint arXiv:2303.12712}, 2023.

\bibitem[Carlini et~al.(2021)Carlini, Tramer, Wallace, Jagielski, Herbert-Voss, Lee, Roberts, Brown, Song, Erlingsson, et~al.]{carlini2021extracting}
Nicholas Carlini, Florian Tramer, Eric Wallace, Matthew Jagielski, Ariel Herbert-Voss, Katherine Lee, Adam Roberts, Tom Brown, Dawn Song, Ulfar Erlingsson, et~al.
\newblock Extracting training data from large language models.
\newblock In \emph{30th USENIX Security Symposium (USENIX Security 21)}, pages 2633--2650, 2021.

\bibitem[Carlini et~al.(2022)Carlini, Jagielski, Zhang, Papernot, Terzis, and Tramer]{carlini2022privacy}
Nicholas Carlini, Matthew Jagielski, Chiyuan Zhang, Nicolas Papernot, Andreas Terzis, and Florian Tramer.
\newblock The privacy onion effect: Memorization is relative.
\newblock \emph{Advances in Neural Information Processing Systems}, 35:\penalty0 13263--13276, 2022.

\bibitem[Chen et~al.(2021)Chen, Tworek, Jun, Yuan, de~Oliveira~Pinto, Kaplan, Edwards, Burda, Joseph, Brockman, Ray, Puri, Krueger, Petrov, Khlaaf, Sastry, Mishkin, Chan, Gray, Ryder, Pavlov, Power, Kaiser, Bavarian, Winter, Tillet, Such, Cummings, Plappert, Chantzis, Barnes, Herbert-Voss, Guss, Nichol, Paino, Tezak, Tang, Babuschkin, Balaji, Jain, Saunders, Hesse, Carr, Leike, Achiam, Misra, Morikawa, Radford, Knight, Brundage, Murati, Mayer, Welinder, McGrew, Amodei, McCandlish, Sutskever, and Zaremba]{chen2021codex}
Mark Chen, Jerry Tworek, Heewoo Jun, Qiming Yuan, Henrique~Ponde de~Oliveira~Pinto, Jared Kaplan, Harri Edwards, Yuri Burda, Nicholas Joseph, Greg Brockman, Alex Ray, Raul Puri, Gretchen Krueger, Michael Petrov, Heidy Khlaaf, Girish Sastry, Pamela Mishkin, Brooke Chan, Scott Gray, Nick Ryder, Mikhail Pavlov, Alethea Power, Lukasz Kaiser, Mohammad Bavarian, Clemens Winter, Philippe Tillet, Felipe~Petroski Such, Dave Cummings, Matthias Plappert, Fotios Chantzis, Elizabeth Barnes, Ariel Herbert-Voss, William~Hebgen Guss, Alex Nichol, Alex Paino, Nikolas Tezak, Jie Tang, Igor Babuschkin, Suchir Balaji, Shantanu Jain, William Saunders, Christopher Hesse, Andrew~N. Carr, Jan Leike, Josh Achiam, Vedant Misra, Evan Morikawa, Alec Radford, Matthew Knight, Miles Brundage, Mira Murati, Katie Mayer, Peter Welinder, Bob McGrew, Dario Amodei, Sam McCandlish, Ilya Sutskever, and Wojciech Zaremba.
\newblock Evaluating large language models trained on code.
\newblock 2021.

\bibitem[Chen et~al.(2024)Chen, Asai, Mireshghallah, Min, Grimmelmann, Choi, Hajishirzi, Zettlemoyer, and Koh]{chen2024copybench}
Tong Chen, Akari Asai, Niloofar Mireshghallah, Sewon Min, James Grimmelmann, Yejin Choi, Hannaneh Hajishirzi, Luke Zettlemoyer, and Pang~Wei Koh.
\newblock Copybench: Measuring literal and non-literal reproduction of copyright-protected text in language model generation.
\newblock \emph{arXiv preprint arXiv:2407.07087}, 2024.

\bibitem[Cobbe et~al.(2021{\natexlab{a}})Cobbe, Kosaraju, Bavarian, Chen, Jun, Kaiser, Plappert, Tworek, Hilton, Nakano, Hesse, and Schulman]{cobbe2021gsm8k}
Karl Cobbe, Vineet Kosaraju, Mohammad Bavarian, Mark Chen, Heewoo Jun, Lukasz Kaiser, Matthias Plappert, Jerry Tworek, Jacob Hilton, Reiichiro Nakano, Christopher Hesse, and John Schulman.
\newblock Training verifiers to solve math word problems.
\newblock \emph{arXiv preprint arXiv:2110.14168}, 2021{\natexlab{a}}.

\bibitem[Cobbe et~al.(2021{\natexlab{b}})Cobbe, Kosaraju, Bavarian, Chen, Jun, Kaiser, Plappert, Tworek, Hilton, Nakano, et~al.]{cobbe2021training}
Karl Cobbe, Vineet Kosaraju, Mohammad Bavarian, Mark Chen, Heewoo Jun, Lukasz Kaiser, Matthias Plappert, Jerry Tworek, Jacob Hilton, Reiichiro Nakano, et~al.
\newblock Training verifiers to solve math word problems.
\newblock \emph{arXiv preprint arXiv:2110.14168}, 2021{\natexlab{b}}.

\bibitem[DeepSeek-AI et~al.(2025)DeepSeek-AI, Guo, Yang, Zhang, Song, Zhang, Xu, Zhu, Ma, Wang, Bi, Zhang, Yu, Wu, Wu, Gou, Shao, Li, Gao, Liu, Xue, Wang, Wu, Feng, Lu, Zhao, Deng, Zhang, Ruan, Dai, Chen, Ji, Li, Lin, Dai, Luo, Hao, Chen, Li, Zhang, Bao, Xu, Wang, Ding, Xin, Gao, Qu, Li, Guo, Li, Wang, Chen, Yuan, Qiu, Li, Cai, Ni, Liang, Chen, Dong, Hu, Gao, Guan, Huang, Yu, Wang, Zhang, Zhao, Wang, Zhang, Xu, Xia, Zhang, Zhang, Tang, Li, Wang, Li, Tian, Huang, Zhang, Wang, Chen, Du, Ge, Zhang, Pan, Wang, Chen, Jin, Chen, Lu, Zhou, Chen, Ye, Wang, Yu, Zhou, Pan, Li, Zhou, Wu, Ye, Yun, Pei, Sun, Wang, Zeng, Zhao, Liu, Liang, Gao, Yu, Zhang, Xiao, An, Liu, Wang, Chen, Nie, Cheng, Liu, Xie, Liu, Yang, Li, Su, Lin, Li, Jin, Shen, Chen, Sun, Wang, Song, Zhou, Wang, Shan, Li, Wang, Wei, Zhang, Xu, Li, Zhao, Sun, Wang, Yu, Zhang, Shi, Xiong, He, Piao, Wang, Tan, Ma, Liu, Guo, Ou, Wang, Gong, Zou, He, Xiong, Luo, You, Liu, Zhou, Zhu, Xu, Huang, Li, Zheng, Zhu, Ma, Tang, Zha, Yan, Ren, Ren, Sha, Fu, Xu, Xie, Zhang,
  Hao, Ma, Yan, Wu, Gu, Zhu, Liu, Li, Xie, Song, Pan, Huang, Xu, Zhang, and Zhang]{deepseekai2025deepseekr1incentivizingreasoningcapability}
DeepSeek-AI, Daya Guo, Dejian Yang, Haowei Zhang, Junxiao Song, Ruoyu Zhang, Runxin Xu, Qihao Zhu, Shirong Ma, Peiyi Wang, Xiao Bi, Xiaokang Zhang, Xingkai Yu, Yu~Wu, Z.~F. Wu, Zhibin Gou, Zhihong Shao, Zhuoshu Li, Ziyi Gao, Aixin Liu, Bing Xue, Bingxuan Wang, Bochao Wu, Bei Feng, Chengda Lu, Chenggang Zhao, Chengqi Deng, Chenyu Zhang, Chong Ruan, Damai Dai, Deli Chen, Dongjie Ji, Erhang Li, Fangyun Lin, Fucong Dai, Fuli Luo, Guangbo Hao, Guanting Chen, Guowei Li, H.~Zhang, Han Bao, Hanwei Xu, Haocheng Wang, Honghui Ding, Huajian Xin, Huazuo Gao, Hui Qu, Hui Li, Jianzhong Guo, Jiashi Li, Jiawei Wang, Jingchang Chen, Jingyang Yuan, Junjie Qiu, Junlong Li, J.~L. Cai, Jiaqi Ni, Jian Liang, Jin Chen, Kai Dong, Kai Hu, Kaige Gao, Kang Guan, Kexin Huang, Kuai Yu, Lean Wang, Lecong Zhang, Liang Zhao, Litong Wang, Liyue Zhang, Lei Xu, Leyi Xia, Mingchuan Zhang, Minghua Zhang, Minghui Tang, Meng Li, Miaojun Wang, Mingming Li, Ning Tian, Panpan Huang, Peng Zhang, Qiancheng Wang, Qinyu Chen, Qiushi Du, Ruiqi Ge, Ruisong
  Zhang, Ruizhe Pan, Runji Wang, R.~J. Chen, R.~L. Jin, Ruyi Chen, Shanghao Lu, Shangyan Zhou, Shanhuang Chen, Shengfeng Ye, Shiyu Wang, Shuiping Yu, Shunfeng Zhou, Shuting Pan, S.~S. Li, Shuang Zhou, Shaoqing Wu, Shengfeng Ye, Tao Yun, Tian Pei, Tianyu Sun, T.~Wang, Wangding Zeng, Wanjia Zhao, Wen Liu, Wenfeng Liang, Wenjun Gao, Wenqin Yu, Wentao Zhang, W.~L. Xiao, Wei An, Xiaodong Liu, Xiaohan Wang, Xiaokang Chen, Xiaotao Nie, Xin Cheng, Xin Liu, Xin Xie, Xingchao Liu, Xinyu Yang, Xinyuan Li, Xuecheng Su, Xuheng Lin, X.~Q. Li, Xiangyue Jin, Xiaojin Shen, Xiaosha Chen, Xiaowen Sun, Xiaoxiang Wang, Xinnan Song, Xinyi Zhou, Xianzu Wang, Xinxia Shan, Y.~K. Li, Y.~Q. Wang, Y.~X. Wei, Yang Zhang, Yanhong Xu, Yao Li, Yao Zhao, Yaofeng Sun, Yaohui Wang, Yi~Yu, Yichao Zhang, Yifan Shi, Yiliang Xiong, Ying He, Yishi Piao, Yisong Wang, Yixuan Tan, Yiyang Ma, Yiyuan Liu, Yongqiang Guo, Yuan Ou, Yuduan Wang, Yue Gong, Yuheng Zou, Yujia He, Yunfan Xiong, Yuxiang Luo, Yuxiang You, Yuxuan Liu, Yuyang Zhou, Y.~X. Zhu,
  Yanhong Xu, Yanping Huang, Yaohui Li, Yi~Zheng, Yuchen Zhu, Yunxian Ma, Ying Tang, Yukun Zha, Yuting Yan, Z.~Z. Ren, Zehui Ren, Zhangli Sha, Zhe Fu, Zhean Xu, Zhenda Xie, Zhengyan Zhang, Zhewen Hao, Zhicheng Ma, Zhigang Yan, Zhiyu Wu, Zihui Gu, Zijia Zhu, Zijun Liu, Zilin Li, Ziwei Xie, Ziyang Song, Zizheng Pan, Zhen Huang, Zhipeng Xu, Zhongyu Zhang, and Zhen Zhang.
\newblock {DeepSeek-R1}: Incentivizing reasoning capability in llms via reinforcement learning, 2025.
\newblock URL \url{https://arxiv.org/abs/2501.12948}.

\bibitem[Dubey et~al.(2024)Dubey, Jauhri, Pandey, Kadian, Al-Dahle, Letman, Mathur, Schelten, Yang, Fan, et~al.]{dubey2024llama}
Abhimanyu Dubey, Abhinav Jauhri, Abhinav Pandey, Abhishek Kadian, Ahmad Al-Dahle, Aiesha Letman, Akhil Mathur, Alan Schelten, Amy Yang, Angela Fan, et~al.
\newblock The llama 3 herd of models.
\newblock \emph{arXiv preprint arXiv:2407.21783}, 2024.

\bibitem[Feldman(2020)]{feldman2020does}
Vitaly Feldman.
\newblock Does learning require memorization? a short tale about a long tail.
\newblock In \emph{Proceedings of the 52nd Annual ACM SIGACT Symposium on Theory of Computing}, pages 954--959, 2020.

\bibitem[Feldman and Zhang(2020)]{feldman2020neural}
Vitaly Feldman and Chiyuan Zhang.
\newblock What neural networks memorize and why: Discovering the long tail via influence estimation.
\newblock \emph{Advances in Neural Information Processing Systems}, 33:\penalty0 2881--2891, 2020.

\bibitem[Gulati et~al.(2024)Gulati, Miranda, Chen, Xia, Fronsdal, de~Moraes~Dumont, and Koyejo]{gulati2024putnamaxiom}
Aryan Gulati, Brando Miranda, Eric Chen, Emily Xia, Kai Fronsdal, Bruno de~Moraes~Dumont, and Sanmi Koyejo.
\newblock Putnam-{AXIOM}: A functional and static benchmark for measuring higher level mathematical reasoning.
\newblock In \emph{The 4th Workshop on Mathematical Reasoning and AI at NeurIPS'24}, 2024.
\newblock URL \url{https://openreview.net/forum?id=YXnwlZe0yf}.

\bibitem[He et~al.(2024)He, Luo, Bai, Hu, Thai, Shen, Hu, Han, Huang, Zhang, Liu, Qi, Liu, and Sun]{he2024olympiadbench}
Chaoqun He, Renjie Luo, Yuzhuo Bai, Shengding Hu, Zhen~Leng Thai, Junhao Shen, Jinyi Hu, Xu~Han, Yujie Huang, Yuxiang Zhang, Jie Liu, Lei Qi, Zhiyuan Liu, and Maosong Sun.
\newblock {OlympiadBench}: A challenging benchmark for promoting agi with olympiad-level bilingual multimodal scientific problems, 2024.

\bibitem[Hendrycks et~al.(2021)Hendrycks, Burns, Kadavath, Arora, Basart, Tang, Song, and Steinhardt]{hendrycksmath2021}
Dan Hendrycks, Collin Burns, Saurav Kadavath, Akul Arora, Steven Basart, Eric Tang, Dawn Song, and Jacob Steinhardt.
\newblock Measuring mathematical problem solving with the math dataset.
\newblock \emph{NeurIPS}, 2021.

\bibitem[Hernandez et~al.(2022)Hernandez, Brown, Conerly, DasSarma, Drain, El-Showk, Elhage, Hatfield-Dodds, Henighan, Hume, et~al.]{hernandez2022scaling}
Danny Hernandez, Tom Brown, Tom Conerly, Nova DasSarma, Dawn Drain, Sheer El-Showk, Nelson Elhage, Zac Hatfield-Dodds, Tom Henighan, Tristan Hume, et~al.
\newblock Scaling laws and interpretability of learning from repeated data.
\newblock \emph{arXiv preprint arXiv:2205.10487}, 2022.

\bibitem[Huang et~al.(2023)Huang, Gupta, Zhong, Li, and Chen]{huang2023privacy}
Yangsibo Huang, Samyak Gupta, Zexuan Zhong, Kai Li, and Danqi Chen.
\newblock Privacy implications of retrieval-based language models.
\newblock \emph{arXiv preprint arXiv:2305.14888}, 2023.

\bibitem[jylin04 et~al.()jylin04, JackS, Karvonen, and Can]{OthelloGPT}
jylin04, JackS, Adam Karvonen, and Can.
\newblock Othellogpt learned a bag of heuristics.
\newblock \url{https://www.lesswrong.com/posts/gcpNuEZnxAPayaKBY/othellogpt-learned-a-bag-of-heuristics-1}.
\newblock Accessed on Date (2025-01-28).

\bibitem[Karamolegkou et~al.(2023)Karamolegkou, Li, Zhou, and S{\o}gaard]{karamolegkou2023copyright}
Antonia Karamolegkou, Jiaang Li, Li~Zhou, and Anders S{\o}gaard.
\newblock Copyright violations and large language models.
\newblock \emph{arXiv preprint arXiv:2310.13771}, 2023.

\bibitem[Kojima et~al.(2022)Kojima, Gu, Reid, Matsuo, and Iwasawa]{kojima2022large}
Takeshi Kojima, Shixiang~Shane Gu, Machel Reid, Yutaka Matsuo, and Yusuke Iwasawa.
\newblock Large language models are zero-shot reasoners.
\newblock \emph{Advances in neural information processing systems}, 35:\penalty0 22199--22213, 2022.

\bibitem[Lee et~al.(2021)Lee, Ippolito, Nystrom, Zhang, Eck, Callison-Burch, and Carlini]{lee2021deduplicating}
Katherine Lee, Daphne Ippolito, Andrew Nystrom, Chiyuan Zhang, Douglas Eck, Chris Callison-Burch, and Nicholas Carlini.
\newblock Deduplicating training data makes language models better.
\newblock \emph{arXiv preprint arXiv:2107.06499}, 2021.

\bibitem[Li et~al.(2024{\natexlab{a}})Li, Beeching, Tunstall, Lipkin, Soletskyi, Huang, Rasul, Yu, Jiang, Shen, Qin, Dong, Zhou, Fleureau, Lample, and Polu]{numina_math_datasets}
Jia Li, Edward Beeching, Lewis Tunstall, Ben Lipkin, Roman Soletskyi, Shengyi~Costa Huang, Kashif Rasul, Longhui Yu, Albert Jiang, Ziju Shen, Zihan Qin, Bin Dong, Li~Zhou, Yann Fleureau, Guillaume Lample, and Stanislas Polu.
\newblock Numinamath.
\newblock \url{[https://github.com/project-numina/aimo-progress-prize](https://github.com/project-numina/aimo-progress-prize/blob/main/report/numina_dataset.pdf)}, 2024{\natexlab{a}}.

\bibitem[Li et~al.(2024{\natexlab{b}})Li, Cui, Zhao, Kong, and Bi]{li2024gsm}
Qintong Li, Leyang Cui, Xueliang Zhao, Lingpeng Kong, and Wei Bi.
\newblock Gsm-plus: A comprehensive benchmark for evaluating the robustness of llms as mathematical problem solvers.
\newblock \emph{arXiv preprint arXiv:2402.19255}, 2024{\natexlab{b}}.

\bibitem[Lightman et~al.(2023)Lightman, Kosaraju, Burda, Edwards, Baker, Lee, Leike, Schulman, Sutskever, and Cobbe]{lightman2023let}
Hunter Lightman, Vineet Kosaraju, Yura Burda, Harri Edwards, Bowen Baker, Teddy Lee, Jan Leike, John Schulman, Ilya Sutskever, and Karl Cobbe.
\newblock Let's verify step by step.
\newblock \emph{arXiv preprint arXiv:2305.20050}, 2023.

\bibitem[Liu et~al.(2024)Liu, Xia, Wang, and Zhang]{liu2024your}
Jiawei Liu, Chunqiu~Steven Xia, Yuyao Wang, and Lingming Zhang.
\newblock Is your code generated by chatgpt really correct? rigorous evaluation of large language models for code generation.
\newblock \emph{Advances in Neural Information Processing Systems}, 36, 2024.

\bibitem[Mirzadeh et~al.(2024)Mirzadeh, Alizadeh, Shahrokhi, Tuzel, Bengio, and Farajtabar]{mirzadeh2024gsm}
Iman Mirzadeh, Keivan Alizadeh, Hooman Shahrokhi, Oncel Tuzel, Samy Bengio, and Mehrdad Farajtabar.
\newblock Gsm-symbolic: Understanding the limitations of mathematical reasoning in large language models.
\newblock \emph{arXiv preprint arXiv:2410.05229}, 2024.

\bibitem[Nasr et~al.(2023)Nasr, Carlini, Hayase, Jagielski, Cooper, Ippolito, Choquette-Choo, Wallace, Tram{\`e}r, and Lee]{nasr2023scalable}
Milad Nasr, Nicholas Carlini, Jonathan Hayase, Matthew Jagielski, A~Feder Cooper, Daphne Ippolito, Christopher~A Choquette-Choo, Eric Wallace, Florian Tram{\`e}r, and Katherine Lee.
\newblock Scalable extraction of training data from (production) language models.
\newblock \emph{arXiv preprint arXiv:2311.17035}, 2023.

\bibitem[Nikankin et~al.(2024)Nikankin, Reusch, Mueller, and Belinkov]{nikankin2024arithmetic}
Yaniv Nikankin, Anja Reusch, Aaron Mueller, and Yonatan Belinkov.
\newblock Arithmetic without algorithms: Language models solve math with a bag of heuristics.
\newblock \emph{arXiv preprint arXiv:2410.21272}, 2024.

\bibitem[OpenAI(2024)]{openaio1}
OpenAI.
\newblock {OpenAI o1}.
\newblock 2024.
\newblock URL \url{https://openai.com/index/openai-o1-system-card/}.

\bibitem[Patel et~al.(2021)Patel, Bhattamishra, and Goyal]{patel-etal-2021-nlp}
Arkil Patel, Satwik Bhattamishra, and Navin Goyal.
\newblock Are {NLP} models really able to solve simple math word problems?
\newblock In Kristina Toutanova, Anna Rumshisky, Luke Zettlemoyer, Dilek Hakkani-Tur, Iz~Beltagy, Steven Bethard, Ryan Cotterell, Tanmoy Chakraborty, and Yichao Zhou, editors, \emph{Proceedings of the 2021 Conference of the North American Chapter of the Association for Computational Linguistics: Human Language Technologies}, pages 2080--2094, Online, June 2021. Association for Computational Linguistics.
\newblock \doi{10.18653/v1/2021.naacl-main.168}.
\newblock URL \url{https://aclanthology.org/2021.naacl-main.168/}.

\bibitem[Rein et~al.(2023)Rein, Hou, Stickland, Petty, Pang, Dirani, Michael, and Bowman]{rein2023gpqa}
David Rein, Betty~Li Hou, Asa~Cooper Stickland, Jackson Petty, Richard~Yuanzhe Pang, Julien Dirani, Julian Michael, and Samuel~R Bowman.
\newblock Gpqa: A graduate-level google-proof q\&a benchmark.
\newblock \emph{arXiv preprint arXiv:2311.12022}, 2023.

\bibitem[Shah et~al.(2024)Shah, Yu, Lyu, Park, Yu, He, Ke, Mozer, Bengio, Arora, et~al.]{shah2024ai}
Vedant Shah, Dingli Yu, Kaifeng Lyu, Simon Park, Jiatong Yu, Yinghui He, Nan~Rosemary Ke, Michael Mozer, Yoshua Bengio, Sanjeev Arora, et~al.
\newblock Ai-assisted generation of difficult math questions.
\newblock \emph{arXiv preprint arXiv:2407.21009}, 2024.

\bibitem[Shao et~al.(2024)Shao, Wang, Zhu, Xu, Song, Bi, Zhang, Zhang, Li, Wu, et~al.]{shao2024deepseekmath}
Zhihong Shao, Peiyi Wang, Qihao Zhu, Runxin Xu, Junxiao Song, Xiao Bi, Haowei Zhang, Mingchuan Zhang, YK~Li, Y~Wu, et~al.
\newblock Deepseekmath: Pushing the limits of mathematical reasoning in open language models.
\newblock \emph{arXiv preprint arXiv:2402.03300}, 2024.

\bibitem[Shi et~al.(2023{\natexlab{a}})Shi, Chen, Misra, Scales, Dohan, Chi, Sch{\"a}rli, and Zhou]{shi2023large}
Freda Shi, Xinyun Chen, Kanishka Misra, Nathan Scales, David Dohan, Ed~H Chi, Nathanael Sch{\"a}rli, and Denny Zhou.
\newblock Large language models can be easily distracted by irrelevant context.
\newblock In \emph{International Conference on Machine Learning}, pages 31210--31227. PMLR, 2023{\natexlab{a}}.

\bibitem[Shi et~al.(2023{\natexlab{b}})Shi, Ajith, Xia, Huang, Liu, Blevins, Chen, and Zettlemoyer]{shi2023detecting}
Weijia Shi, Anirudh Ajith, Mengzhou Xia, Yangsibo Huang, Daogao Liu, Terra Blevins, Danqi Chen, and Luke Zettlemoyer.
\newblock Detecting pretraining data from large language models.
\newblock \emph{arXiv preprint arXiv:2310.16789}, 2023{\natexlab{b}}.

\bibitem[Srivastava et~al.(2024)Srivastava, PV, Menon, Sukumar, Philipose, Prince, Thomas, et~al.]{srivastava2024functional}
Saurabh Srivastava, Anto PV, Shashank Menon, Ajay Sukumar, Alan Philipose, Stevin Prince, Sooraj Thomas, et~al.
\newblock Functional benchmarks for robust evaluation of reasoning performance, and the reasoning gap.
\newblock \emph{arXiv preprint arXiv:2402.19450}, 2024.

\bibitem[Team et~al.(2024{\natexlab{a}})Team, Georgiev, Lei, Burnell, Bai, Gulati, Tanzer, Vincent, Pan, Wang, et~al.]{team2024gemini}
Gemini Team, Petko Georgiev, Ving~Ian Lei, Ryan Burnell, Libin Bai, Anmol Gulati, Garrett Tanzer, Damien Vincent, Zhufeng Pan, Shibo Wang, et~al.
\newblock Gemini 1.5: Unlocking multimodal understanding across millions of tokens of context.
\newblock \emph{arXiv preprint arXiv:2403.05530}, 2024{\natexlab{a}}.

\bibitem[Team et~al.(2024{\natexlab{b}})Team, Riviere, Pathak, Sessa, Hardin, Bhupatiraju, Hussenot, Mesnard, Shahriari, Ram{\'e}, et~al.]{team2024gemma}
Gemma Team, Morgane Riviere, Shreya Pathak, Pier~Giuseppe Sessa, Cassidy Hardin, Surya Bhupatiraju, L{\'e}onard Hussenot, Thomas Mesnard, Bobak Shahriari, Alexandre Ram{\'e}, et~al.
\newblock Gemma 2: Improving open language models at a practical size.
\newblock \emph{arXiv preprint arXiv:2408.00118}, 2024{\natexlab{b}}.

\bibitem[Team(2024)]{qwq-32b-preview}
Qwen Team.
\newblock {QwQ}: Reflect deeply on the boundaries of the unknown, November 2024.
\newblock URL \url{https://qwenlm.github.io/blog/qwq-32b-preview/}.

\bibitem[Tirumala et~al.(2022)Tirumala, Markosyan, Zettlemoyer, and Aghajanyan]{tirumala2022memorization}
Kushal Tirumala, Aram Markosyan, Luke Zettlemoyer, and Armen Aghajanyan.
\newblock Memorization without overfitting: Analyzing the training dynamics of large language models.
\newblock \emph{Advances in Neural Information Processing Systems}, 35:\penalty0 38274--38290, 2022.

\bibitem[Wang et~al.(2022)Wang, Wei, Schuurmans, Le, Chi, Narang, Chowdhery, and Zhou]{wang2022self}
Xuezhi Wang, Jason Wei, Dale Schuurmans, Quoc Le, Ed~Chi, Sharan Narang, Aakanksha Chowdhery, and Denny Zhou.
\newblock Self-consistency improves chain of thought reasoning in language models.
\newblock \emph{arXiv preprint arXiv:2203.11171}, 2022.

\bibitem[Wang et~al.(2024)Wang, Ma, Zhang, Ni, Chandra, Guo, Ren, Arulraj, He, Jiang, et~al.]{wang2024mmlu}
Yubo Wang, Xueguang Ma, Ge~Zhang, Yuansheng Ni, Abhranil Chandra, Shiguang Guo, Weiming Ren, Aaran Arulraj, Xuan He, Ziyan Jiang, et~al.
\newblock Mmlu-pro: A more robust and challenging multi-task language understanding benchmark.
\newblock \emph{arXiv preprint arXiv:2406.01574}, 2024.

\bibitem[Wei et~al.(2024)Wei, Shi, Huang, Smith, Zhang, Zettlemoyer, Li, and Henderson]{wei2024evaluating}
Boyi Wei, Weijia Shi, Yangsibo Huang, Noah~A Smith, Chiyuan Zhang, Luke Zettlemoyer, Kai Li, and Peter Henderson.
\newblock Evaluating copyright takedown methods for language models.
\newblock \emph{arXiv preprint arXiv:2406.18664}, 2024.

\bibitem[Wei et~al.(2022)Wei, Wang, Schuurmans, Bosma, Xia, Chi, Le, Zhou, et~al.]{wei2022chain}
Jason Wei, Xuezhi Wang, Dale Schuurmans, Maarten Bosma, Fei Xia, Ed~Chi, Quoc~V Le, Denny Zhou, et~al.
\newblock Chain-of-thought prompting elicits reasoning in large language models.
\newblock \emph{Advances in neural information processing systems}, 35:\penalty0 24824--24837, 2022.

\bibitem[Wu et~al.(2024)Wu, Sun, Li, Welleck, and Yang]{wu2024empirical}
Yangzhen Wu, Zhiqing Sun, Shanda Li, Sean Welleck, and Yiming Yang.
\newblock An empirical analysis of compute-optimal inference for problem-solving with language models.
\newblock 2024.

\bibitem[Wu et~al.(2023)Wu, Qiu, Ross, Aky{\"u}rek, Chen, Wang, Kim, Andreas, and Kim]{wu2023reasoning}
Zhaofeng Wu, Linlu Qiu, Alexis Ross, Ekin Aky{\"u}rek, Boyuan Chen, Bailin Wang, Najoung Kim, Jacob Andreas, and Yoon Kim.
\newblock Reasoning or reciting? exploring the capabilities and limitations of language models through counterfactual tasks.
\newblock \emph{arXiv preprint arXiv:2307.02477}, 2023.

\bibitem[Xie et~al.(2024)Xie, Huang, Zhang, Yu, Chen, Lin, Li, Ghazi, and Kumar]{xie2024memorization}
Chulin Xie, Yangsibo Huang, Chiyuan Zhang, Da~Yu, Xinyun Chen, Bill~Yuchen Lin, Bo~Li, Badih Ghazi, and Ravi Kumar.
\newblock On memorization of large language models in logical reasoning.
\newblock \emph{arXiv preprint arXiv:2410.23123}, 2024.

\bibitem[Yan et~al.(2024)Yan, Mao, Ji, Zhang, Patil, Stoica, and Gonzalez]{berkeley-function-calling-leaderboard}
Fanjia Yan, Huanzhi Mao, Charlie Cheng-Jie Ji, Tianjun Zhang, Shishir~G. Patil, Ion Stoica, and Joseph~E. Gonzalez.
\newblock Berkeley function calling leaderboard.
\newblock \url{https://gorilla.cs.berkeley.edu/blogs/8_berkeley_function_calling_leaderboard.html}, 2024.

\bibitem[Yang et~al.(2024)Yang, Zhang, Hui, Gao, Yu, Li, Liu, Tu, Zhou, Lin, et~al.]{yang2024qwen25}
An~Yang, Beichen Zhang, Binyuan Hui, Bofei Gao, Bowen Yu, Chengpeng Li, Dayiheng Liu, Jianhong Tu, Jingren Zhou, Junyang Lin, et~al.
\newblock Qwen2. 5-math technical report: Toward mathematical expert model via self-improvement.
\newblock \emph{arXiv preprint arXiv:2409.12122}, 2024.

\bibitem[Yu et~al.(2024)Yu, Jiang, Shi, YU, Liu, Zhang, Kwok, Li, Weller, and Liu]{yu2024metamath}
Longhui Yu, Weisen Jiang, Han Shi, Jincheng YU, Zhengying Liu, Yu~Zhang, James Kwok, Zhenguo Li, Adrian Weller, and Weiyang Liu.
\newblock Metamath: Bootstrap your own mathematical questions for large language models.
\newblock In \emph{The Twelfth International Conference on Learning Representations}, 2024.
\newblock URL \url{https://openreview.net/forum?id=N8N0hgNDRt}.

\bibitem[Yue et~al.(2024)Yue, Zheng, Zhang, and Chen]{yue2024mammoth2}
Xiang Yue, Tuney Zheng, Ge~Zhang, and Wenhu Chen.
\newblock Mammoth2: Scaling instructions from the web.
\newblock \emph{Advances in Neural Information Processing Systems}, 2024.

\bibitem[Zhang et~al.(2021)Zhang, Bengio, Hardt, Recht, and Vinyals]{zhang2021understanding}
Chiyuan Zhang, Samy Bengio, Moritz Hardt, Benjamin Recht, and Oriol Vinyals.
\newblock Understanding deep learning (still) requires rethinking generalization.
\newblock \emph{Communications of the ACM}, 64\penalty0 (3):\penalty0 107--115, 2021.

\bibitem[Zhang et~al.(2023)Zhang, Ippolito, Lee, Jagielski, Tram{\`e}r, and Carlini]{zhang2023counterfactual}
Chiyuan Zhang, Daphne Ippolito, Katherine Lee, Matthew Jagielski, Florian Tram{\`e}r, and Nicholas Carlini.
\newblock Counterfactual memorization in neural language models.
\newblock \emph{Advances in Neural Information Processing Systems}, 36:\penalty0 39321--39362, 2023.

\bibitem[Zhang et~al.(2024)Zhang, Da, Lee, Robinson, Wu, Song, Zhao, Raja, Slack, Lyu, et~al.]{zhang2024careful}
Hugh Zhang, Jeff Da, Dean Lee, Vaughn Robinson, Catherine Wu, Will Song, Tiffany Zhao, Pranav Raja, Dylan Slack, Qin Lyu, et~al.
\newblock A careful examination of large language model performance on grade school arithmetic.
\newblock \emph{arXiv preprint arXiv:2405.00332}, 2024.

\bibitem[Zheng et~al.(2023)Zheng, Zhou, Meng, Zhou, and Huang]{zheng2023large}
Chujie Zheng, Hao Zhou, Fandong Meng, Jie Zhou, and Minlie Huang.
\newblock On large language models' selection bias in multi-choice questions.
\newblock \emph{arXiv preprint arXiv:2309.03882}, 2023.

\bibitem[Zhou et~al.(2023)Zhou, Lu, Mishra, Brahma, Basu, Luan, Zhou, and Hou]{zhou2023instruction}
Jeffrey Zhou, Tianjian Lu, Swaroop Mishra, Siddhartha Brahma, Sujoy Basu, Yi~Luan, Denny Zhou, and Le~Hou.
\newblock Instruction-following evaluation for large language models.
\newblock \emph{arXiv preprint arXiv:2311.07911}, 2023.

\bibitem[Zou et~al.(2024)Zou, Guo, Yang, Zhang, Hu, and Zhang]{zou2024dynamath}
Chengke Zou, Xingang Guo, Rui Yang, Junyu Zhang, Bin Hu, and Huan Zhang.
\newblock Dynamath: A dynamic visual benchmark for evaluating mathematical reasoning robustness of vision language models.
\newblock \emph{arXiv preprint arXiv:2411.00836}, 2024.

\end{thebibliography}
\bibliographystyle{plainnat}

\newpage
\appendix
\onecolumn

\section{Version Information of the Models}
\label{appendix:model}

We consider the following models in the paper.
\begin{itemize}[itemsep=1pt, parsep=1pt, topsep=1pt]
    \item long-CoT models: o1-preview, o1-mini~\citep{openaio1}, Gemini 2.0 flash thinking 
    \item closed-source models:  GPT-4o, GPT-4 Turbo~\citep{achiam2023gpt}, Gemini 1.5 Pro, Gemini 2.0 flash \citep{team2024gemini}, Claude 3.5 Sonnet, Claude 3 Opus \citep{claude35};
    \item open-sourced general-purpose models: Llama 3.1~\citep{dubey2024llama}, Gemma 2~\citep{team2024gemma}, Phi-3.5~\citep{abdin2024phi3technicalreporthighly};
    \item math-specific models: MetaMath~\citep{yu2024metamath}, MAmmoTH2~\citep{yue2024mammoth2}, Deepseek-Math~\citep{shao2024deepseekmath}, Qwen2.5-Math~\citep{yang2024qwen25}, NuminaMath~\citep{numina_math_datasets}, Mathtral\footnote{Mathtral \url{https://mistral.ai/news/mathstral/} }.
\end{itemize}

\begin{table*}[htbp]
\caption{Version information of the models}
\centering 
\begin{tabular}{lcr}
 \toprule
\textbf{Model} & \textbf{Provider} & \textbf{Version/Link} \\
\midrule
Gemini-2.0-flash-thinking-exp & Google DeepMind & 2024-12-19 \\
o1-preview & OpenAI & 2024-09-12\\
o1-mini & OpenAI & 2024-09-12\\
\midrule
Gemini-2.0-flash-exp & Google DeepMind& 2024-12-11\\
Gemini-1.5-pro & Google DeepMind& gemini-1.5-pro-002 \\
GPT-4o & OpenAI & 2024-08-06 \\
GPT-4-turbo &OpenAI & 2024-04-09\\
Claude-3.5-sonnet & Anthropic& 2024-10-22\\
Claude-3-opus &Anthropic& 2024-02-29\\
\midrule
Llama-3.1-8B-Instruct&  Open-Sourced & \url{https://huggingface.co/meta-llama/Llama-3.1-8B-Instruct}\\
Gemma-2-9b-it & Open-Sourced & \url{https://huggingface.co/google/gemma-2-9b-it}\\
Phi-3.5-mini-instruct &  Open-Sourced& \url{https://huggingface.co/microsoft/Phi-3.5-mini-instruct} \\
\midrule
Deepseek-math-7b-rl & Open-Sourced& \url{https://huggingface.co/deepseek-ai/deepseek-math-7b-rl}\\
Qwen2.5-Math-7B-Instruct & Open-Sourced& \url{https://huggingface.co/Qwen/Qwen2.5-Math-7B-Instruct}\\
Mathstral-7b-v0.1& Open-Sourced& \url{https://huggingface.co/mistralai/Mathstral-7B-v0.1}\\ 
NuminaMath-7B-CoT & Open-Sourced& \url{https://huggingface.co/AI-MO/NuminaMath-7B-CoT} \\
MetaMath-13B-V1.0& Open-Sourced& \url{https://huggingface.co/meta-math/MetaMath-13B-V1.0}\\
MAmmoTH2-8B & Open-Sourced& \url{https://huggingface.co/TIGER-Lab/MAmmoTH2-8B}\\
\bottomrule
\end{tabular}
\label{tab:model}
\end{table*}

\section{Benchmark Statistics}
\label{appendix:benchmark}

\begin{table}[ht]
\centering
\small
\caption{Number of problems corresponding to different subjects.}
\begin{tabular}{l|r}
\toprule
\textbf{Subject} & \textbf{Number (Percentage)} \\ \midrule
Prealgebra & 35 (12.54 \%) \\
Algebra & 79 (28.32 \%) \\ %
Number Theory & 36 (12.90 \%) \\
Counting \& Probability & 38 (13.62 \%) \\
Geometry & 21 (7.53 \%) \\
Intermediate Algebra & 48 (17.20 \%) \\
Precalculus & 22 (7.89 \%) \\\midrule
\textbf{Total} & 279 \\\bottomrule
\end{tabular}
\label{tab:stat}
\end{table}

\section{Additional Experimental Results}
\label{appendix:exp}

\subsection{Categorizing Model Responses Across Problem Variations}
\label{appendix:category}
Recall that for each problem, we have a \SAME modification which can be solved using the same method as the original problem, and a \HARD modification which requires more difficult problem-solving skills. Therefore, there are 8 possible cases regarding the correctness of the model's responses to the three problems. Modulo the fluctuations of the model's correctness among the \SAME variations, we can summarize the model's responses into the following 4 cases:
\begin{itemize}[itemsep=1pt, parsep=1pt, topsep=1pt]
    \item \textbf{Case I}: at least one of the original problem and the \SAME modification is solved \textit{correctly}, and the \HARD modification is also solved \textit{correctly}.
    \item \textbf{Case II}: both the original problem and the \SAME modification are solved \textit{incorrectly}, and the \HARD modification is also solved \textit{incorrectly}.
    \item \textbf{Case III}: both the original problem and the \SAME modification are solved \textit{incorrectly}, but the \HARD modification is solved \textit{correctly}.
    \item \textbf{Case IV}: at least one of the original problem and the \SAME modification is solved \textit{correctly}, but the \HARD modification is solved \textit{incorrectly}.
\end{itemize}
For each of the models, we calculate the percentage of the responses in \cref{tab:cate}. As expected, stronger models have a higher percentage of Case I responses and a lower percentage of Case II responses. Interestingly, the percentages of Case III responses are small (less than 10\%) but non-zero, where the models cannot solve the easier variants but can solve the hard variant correctly. After manual inspection, we found that this is due to the misalignment between the models' capabilities and the annotators' perception of the difficulties of math problems.

\begin{table*}[t]
\caption{Number and percentage of the models' responses that belong to each of the four categories.}
\centering
\resizebox{0.9\textwidth}{!}{
\begin{tabular}{lccccccc}
 \toprule
 Model &  Case I &  Case II & Case III & Case IV &\\ \midrule
Gemini-2.0-flash-thinking-exp &  212 (75.99 \%) & 5 (1.79 \%) & 6 (2.15 \%) & 56 (20.07 \%) &  \\ 
o1-preview &  194 (69.53 \%) & 10 (3.58 \%) & 8 (2.87 \%) & 67 (24.01 \%) &  \\ 
o1-mini &  218 (78.14 \%) & 4 (1.43 \%) & 1 (0.36 \%) & 56 (20.07 \%) &  \\ 
\midrule
Gemini-2.0-flash-exp &  176 (63.08 \%) & 11 (3.94 \%) & 11 (3.94 \%) & 81 (29.03 \%) &  \\ 
Gemini-1.5-pro &  145 (51.97 \%) & 28 (10.04 \%) & 13 (4.66 \%) & 93 (33.33 \%) &  \\ 
GPT-4o &  94 (33.69 \%) & 56 (20.07 \%) & 16 (5.73 \%) & 113 (40.50 \%) &  \\ 
GPT-4-turbo &  81 (29.03 \%) & 72 (25.81 \%) & 15 (5.38 \%) & 111 (39.78 \%) &  \\ 
Claude-3.5-Sonnet &  88 (31.54 \%) & 56 (20.07 \%) & 20 (7.17 \%) & 115 (41.22 \%) &  \\ 
Claude-3-Opus &  49 (17.56 \%) & 99 (35.48 \%) & 25 (8.96 \%) & 106 (37.99 \%) &  \\ 
\midrule
Llama-3.1-8B-Instruct &  21 (7.53 \%) & 137 (49.10 \%) & 7 (2.51 \%) & 114 (40.86 \%) &  \\ 
Gemma-2-9b-it &  22 (7.89 \%) & 164 (58.78 \%) & 11 (3.94 \%) & 82 (29.39 \%) &  \\ 
Phi-3.5-mini-instruct &  22 (7.89 \%) & 161 (57.71 \%) & 18 (6.45 \%) & 78 (27.96 \%) &  \\ 
\midrule
Deepseek-math-7b-rl &  25 (8.96 \%) & 138 (49.46 \%) & 13 (4.66 \%) & 103 (36.92 \%) &  \\ 
Qwen2.5-Math-7B-Instruct &  61 (21.86 \%) & 70 (25.09 \%) & 15 (5.38 \%) & 133 (47.67 \%) &  \\ 
Mathstral-7b-v0.1 &  28 (10.04 \%) & 136 (48.75 \%) & 13 (4.66 \%) & 102 (36.56 \%) &  \\ 
NuminaMath-7B-CoT &  39 (13.98 \%) & 118 (42.29 \%) & 9 (3.23 \%) & 113 (40.50 \%) &  \\ 
MetaMath-13B-V1.0 &  6 (2.15 \%) & 199 (71.33 \%) & 10 (3.58 \%) & 64 (22.94 \%) &  \\ 
MAmmoTH2-8B &  9 (3.23 \%) & 201 (72.04 \%) & 12 (4.30 \%) & 57 (20.43 \%) &  \\
\bottomrule
\end{tabular}
}
\label{tab:cate}
\end{table*}

\subsection{Is Mode Collapse a Problem?}
\label{appendix:naive:memorization}

We provide \cref{tab:naive_memorization} to support \cref{sec:naive:memorization}.

\begin{table*}[htbp]
\caption{The number of errors with answers that match the corresponding original answers. The edit distances are normalized by the length of the responses to the original problems.}
\centering
\resizebox{\textwidth}{!}{
\begin{tabular}{lcccccccccccc}
 \toprule
\multirow{3}{*}{\textbf{Model}}   & \multicolumn{6}{c}{\textbf{\SAME}} & \multicolumn{6}{c}{\textbf{\HARD}}   \\ 
\cmidrule(r){2-7}  \cmidrule(r){8-13}
&  \multicolumn{3}{c}{Num. Errors} & \multicolumn{3}{c}{Normalized Edit Distance} & \multicolumn{3}{c}{Num. Errors} & \multicolumn{3}{c}{Normalized Edit Distance}  \\
\cmidrule(r){2-4} \cmidrule(r){5-7} \cmidrule(r){8-10} \cmidrule(r){11-13}
& $n_{\text{same}}$ & $n_{\text{total}}$ & percentage & min. & avg. & max. & $n_{\text{same}}$ & $n_{\text{total}}$ & percentage & min. & avg. & max. \\
\midrule
 
Gemini-2.0-flash-thinking-exp & 2 & 25 & 8.00 & 0.553 & 0.611 & 0.668  & 10 & 61 & 16.39  &  0.508 & 0.679 & 0.976  \\ 
o1-preview & 1 & 34 & 2.94 & 0.652 & 0.652 & 0.652  & 5 & 77 & 6.49  &  0.729 & 1.07 & 1.89  \\ 
o1-mini & 0 & 14 & 0 & N/A & N/A & N/A  & 9 & 60 & 15.00  &  0.559 & 14.7 & 126.0  \\ 
\midrule
Gemini-2.0-flash-exp & 4 & 48 & 8.33 & 0.644 & 0.82 & 1.09  & 13 & 92 & 14.13  &  0.546 & 1.1 & 1.76  \\ 
Gemini-1.5-pro & 5 & 63 & 7.94 & 0.472 & 0.751 & 1.3  & 11 & 121 & 9.09  &  0.257 & 0.866 & 1.58  \\ 
GPT-4o & 4 & 106 & 3.77 & 0.709 & 0.773 & 0.937  & 14 & 169 & 8.28  &  0.489 & 0.777 & 1.2  \\ 
GPT-4-turbo & 5 & 125 & 4.00 & 0.621 & 0.74 & 0.855  & 17 & 183 & 9.29  &  0.636 & 0.932 & 1.61  \\ 
Claude-3.5-Sonnet & 6 & 116 & 5.17 & 0.509 & 0.729 & 0.83  & 13 & 171 & 7.60  &  0.461 & 0.741 & 1.92  \\ 
Claude-3-Opus & 3 & 162 & 1.85 & 0.355 & 0.485 & 0.614  & 15 & 205 & 7.32  &  0.463 & 0.841 & 1.54  \\ 
\midrule
Llama-3.1-8B-Instruct & 13 & 191 & 6.81 & 0.595 & 0.901 & 1.99  & 18 & 251 & 7.17  &  0.618 & 0.946 & 2.7  \\ 
Gemma-2-9b-it & 3 & 202 & 1.49 & 0.361 & 0.506 & 0.716  & 7 & 246 & 2.85  &  0 & 0.662 & 1.08  \\ 
Phi-3.5-mini-instruct & 8 & 199 & 4.02 & 0.427 & 0.61 & 0.832  & 12 & 239 & 5.02  &  0.289 & 0.754 & 1.69  \\ 
\midrule
Deepseek-math-7b-rl & 9 & 186 & 4.84 & 0.189 & 0.423 & 0.676  & 11 & 241 & 4.56  &  0.121 & 1.5 & 4.24  \\ 
Qwen2.5-Math-7B-Instruct & 6 & 135 & 4.44 & 0.376 & 0.591 & 0.813  & 10 & 203 & 4.93  &  0.273 & 1.01 & 4.91  \\ 
Mathstral-7b-v0.1 & 11 & 178 & 6.18 & 0.0989 & 0.645 & 0.964  & 13 & 238 & 5.46  &  0.105 & 0.586 & 0.984  \\ 
NuminaMath-7B-CoT & 12 & 167 & 7.19 & 0.241 & 0.743 & 1.62  & 14 & 231 & 6.06  &  0.204 & 1.04 & 2.22  \\ 
MetaMath-13B-V1.0 & 13 & 258 & 5.04 & 0.27 & 0.55 & 0.748  & 14 & 263 & 5.32  &  0.509 & 0.982 & 2.83  \\ 
MAmmoTH2-8B & 5 & 229 & 2.18 & 0.00214 & 0.666 & 1.25  & 9 & 258 & 3.49  &  0.708 & 0.822 & 1.04  \\

\bottomrule
\end{tabular}
}
\label{tab:naive_memorization}
\end{table*}

\clearpage
\subsection{The Effect of In-Context Learning}
\label{appendix:ICL}

\begin{figure*}[htbp]
    \centering
    \includegraphics[width=0.33\linewidth]{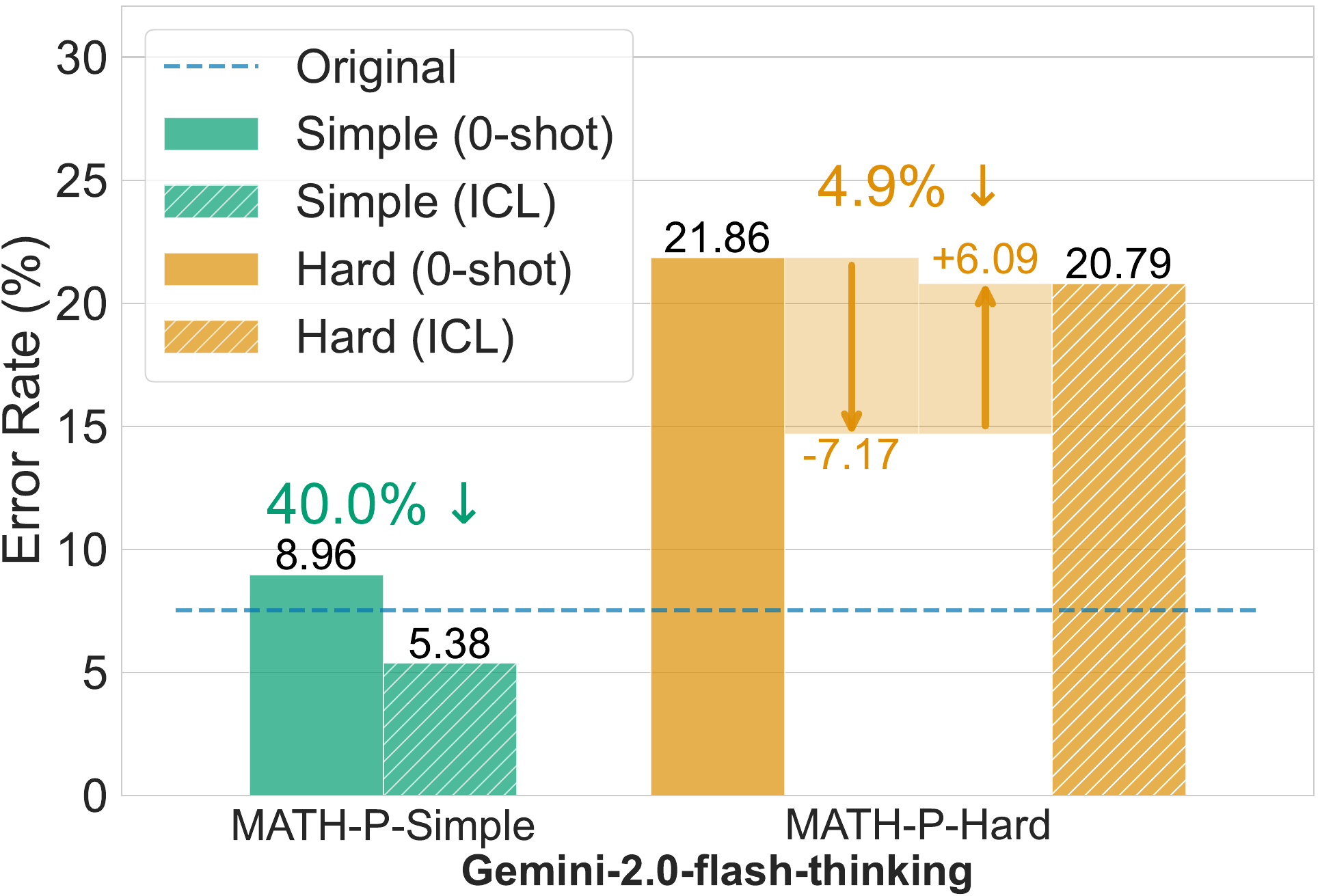}
    \includegraphics[width=0.32\linewidth]{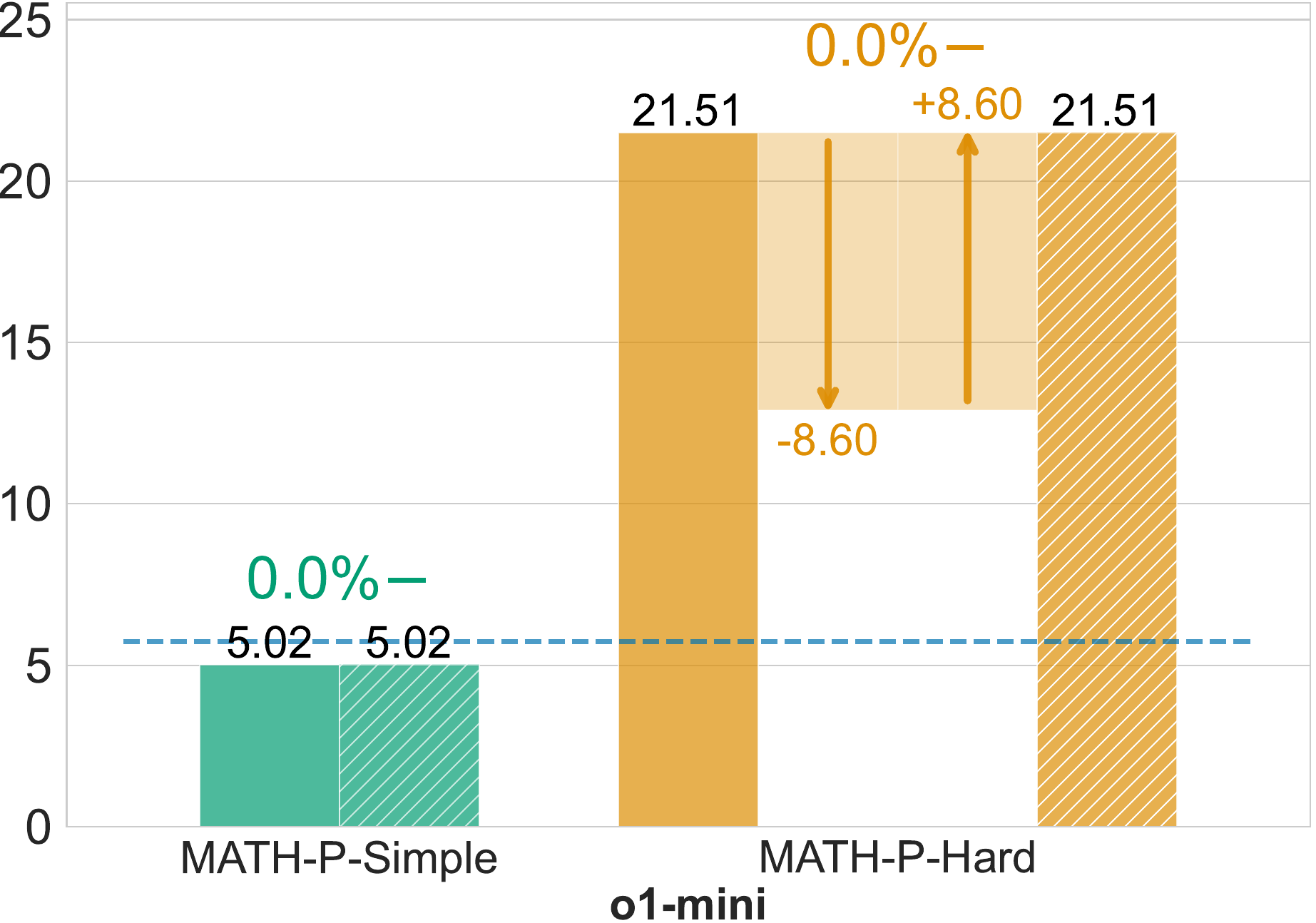}
    \includegraphics[width=0.32\linewidth]{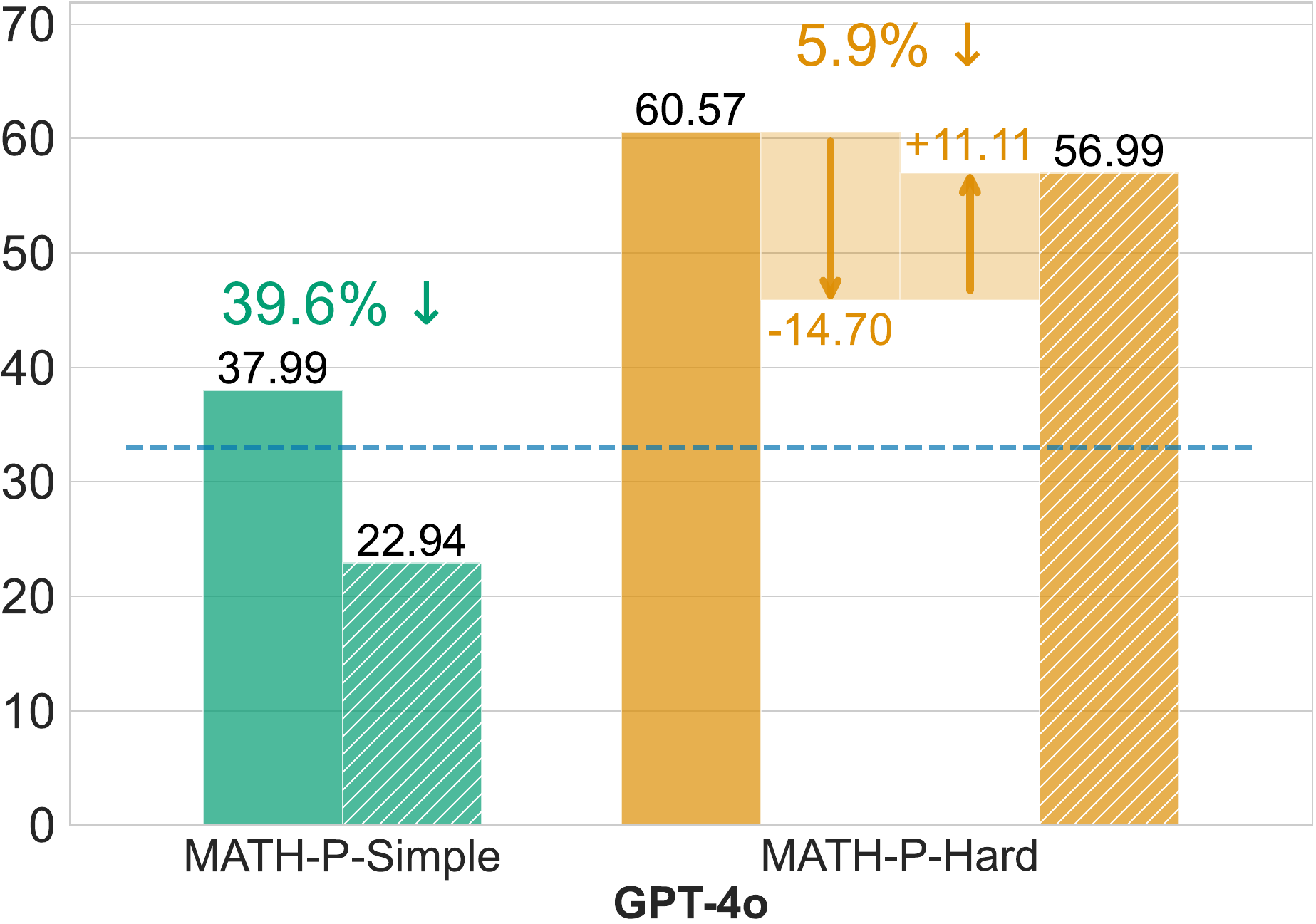}
    \includegraphics[width=0.33\linewidth]{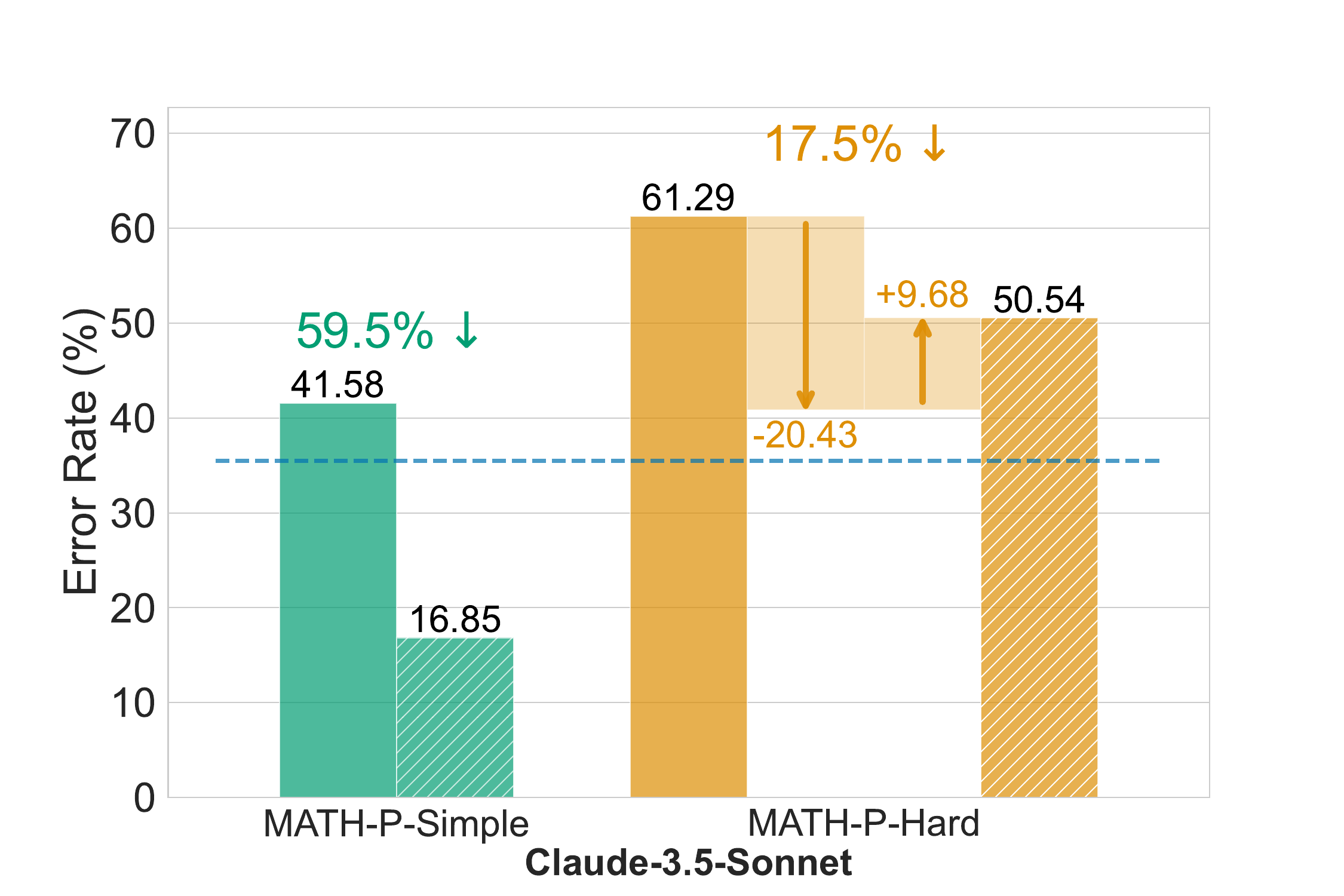}
    \vspace{5mm}
    \includegraphics[width=0.32\linewidth]{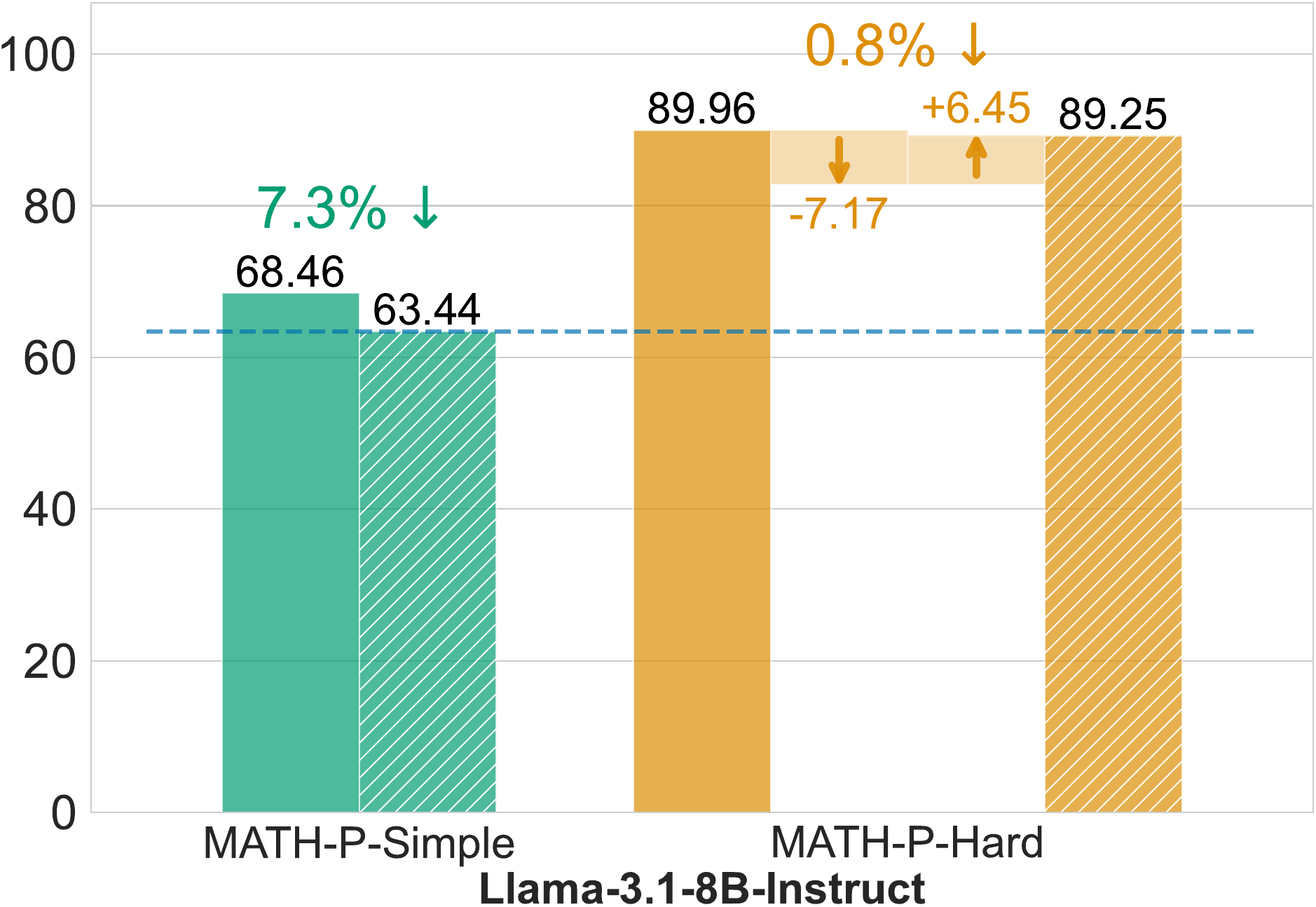}
    \includegraphics[width=0.32\linewidth]{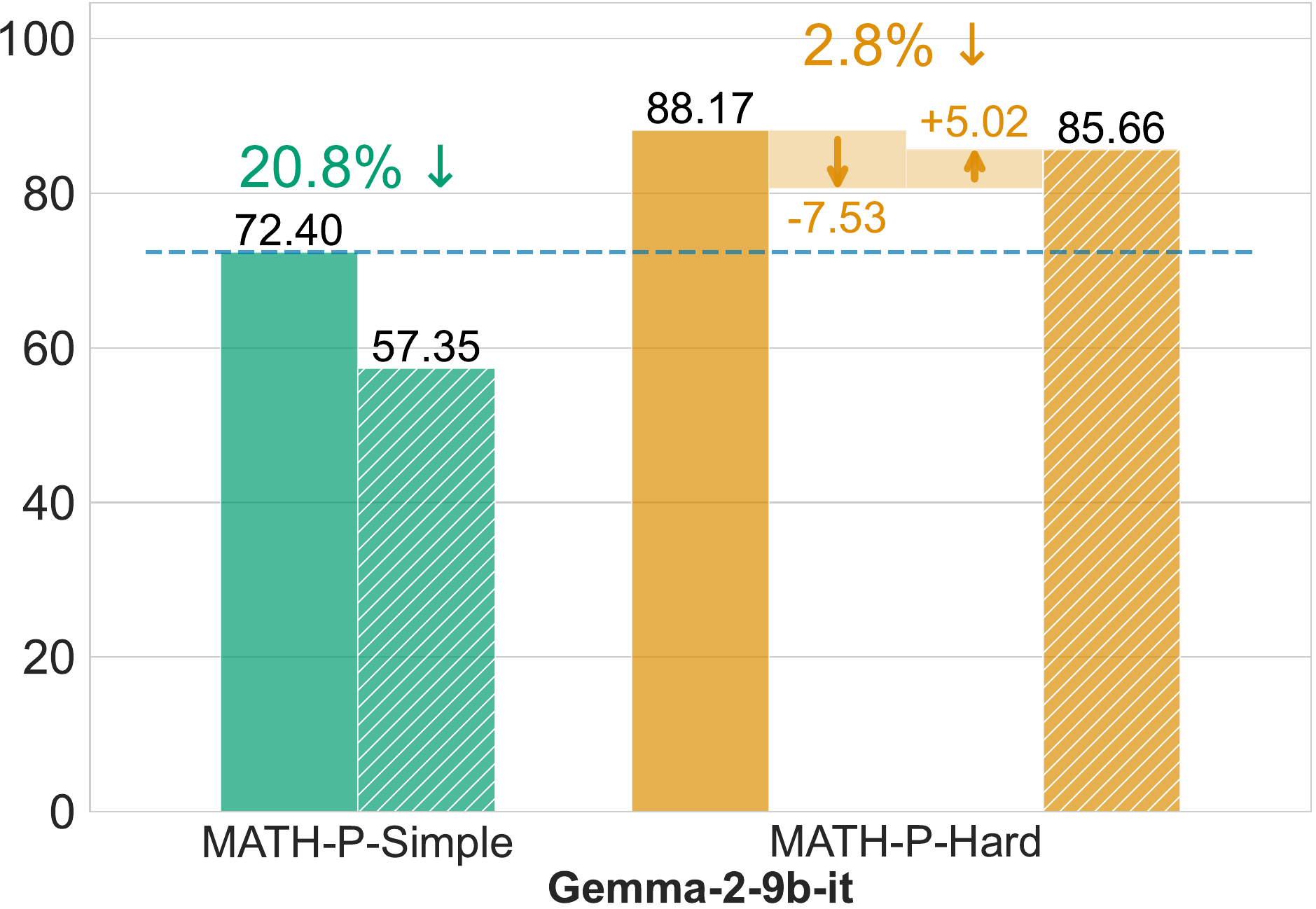}
    \includegraphics[width=0.34\linewidth]{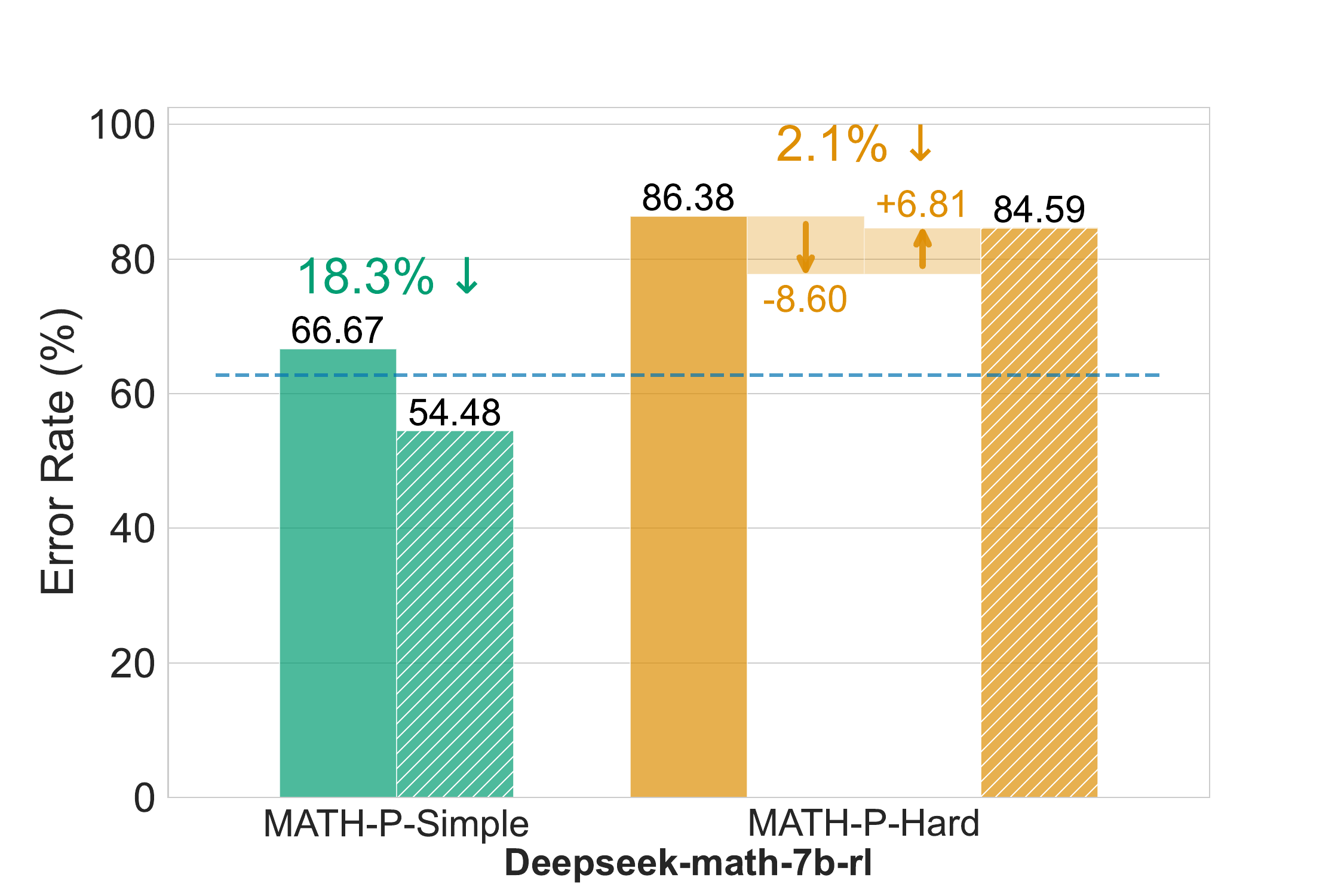}
    \includegraphics[width=0.32\linewidth]{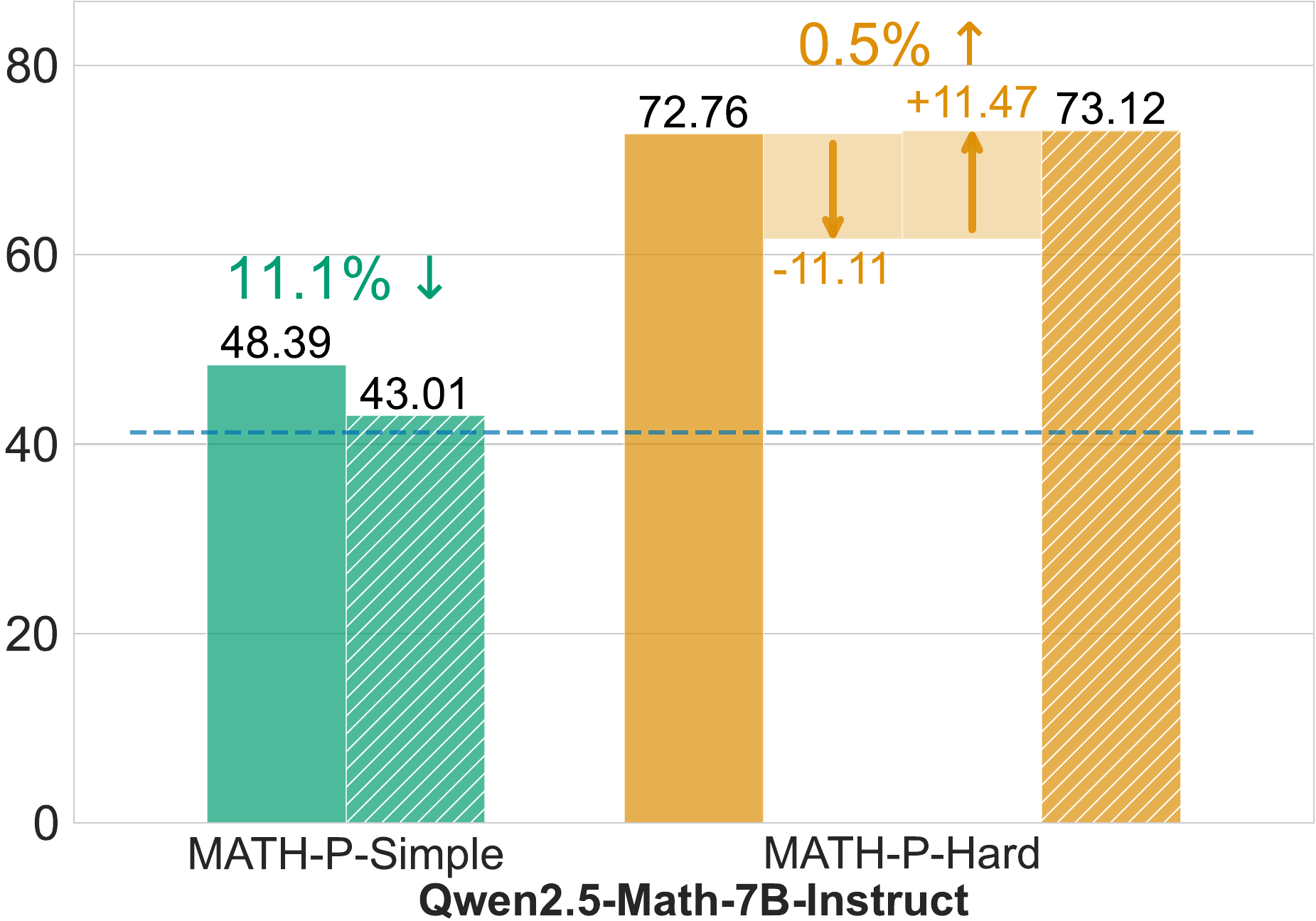}
    
    \caption{The error rates (\%) of the models without and with the original problem and solution as the in-context learning (ICL) example. For \HARD, we decompose the influences of in-context learning into \textbf{ICL effect} (the down arrow $\textcolor{brown}{\boldsymbol{\downarrow}}$), which reduces the error rates, and \textbf{misleading effect} (the up arrow $\textcolor{brown}{\boldsymbol{\uparrow}}$), which increases the error rates.
    }
    \label{fig:icl:full}
\end{figure*}

In \cref{tab:OIC}, we report the performance of in-context learning (ICL) with the corresponding original (unmodified) problem and solution as the in-context learning example. Furthermore, we decompose the influences on \HARD into the \textbf{ICL effect} and the \textbf{misleading effect} in \cref{tab:icl:breakdown} and visualize the influences for representative models in \cref{fig:icl:full}. Please refer to \cref{sec:icl} for the full discussion.

\begin{table*}[htbp]
\caption{Performance comparisons without and with the original problem and solution as the in-context learning example.}
\centering
\resizebox{\textwidth}{!}{
\begin{tabular}{lccccc}
 \toprule
\multirow{2}{*}{\textbf{Model}} & \multirow{2}{*}{\textbf{\Original} (0-shot)}   & \multicolumn{2}{c}{\textbf{\SAME}} & \multicolumn{2}{c}{\textbf{\HARD}}   \\ 
\cmidrule(r){3-4}  \cmidrule(r){5-6}
& & zero-shot & ICL w. original & zero-shot & ICL w. original\\ 
\midrule
Gemini-2.0-flash-thinking-exp & 92.47 & 91.04 & 94.62 & 78.14 & 79.21 \\ 
o1-preview & 87.81 & 87.81 & 91.40 & 72.40 & 74.19 \\ 
o1-mini & 94.27 & 94.98 & 94.98 & 78.49 & 78.49 \\ 
\midrule
Gemini-2.0-flash-exp & 88.17 & 82.80 & 89.96 & 67.03 & 67.38 \\ 
Gemini-1.5-pro & 77.78 & 77.42 & 88.17 & 56.63 & 60.57 \\ 
GPT-4o & 67.03 & 62.01 & 77.06 & 39.43 & 43.01 \\ 
GPT-4-turbo & 56.99 & 55.20 & 69.89 & 34.41 & 39.07 \\ 
Claude-3.5-Sonnet & 64.52 & 58.42 & 83.15 & 38.71 & 49.46 \\ 
Claude-3-Opus & 41.94 & 41.94 & 68.10 & 26.52 & 33.33 \\ 
\midrule
Llama-3.1-8B-Instruct & 36.56 & 31.54 & 36.56 & 10.04 & 10.75 \\ 
Gemma-2-9b-it & 27.60 & 27.60 & 42.65 & 11.83 & 14.34 \\ 
Phi-3.5-mini-instruct & 26.16 & 28.67 & 36.92 & 14.34 & 14.34 \\ 
\midrule
Deepseek-math-7b-rl & 37.28 & 33.33 & 45.52 & 13.62 & 15.41 \\ 
Qwen2.5-Math-7B-Instruct & 58.78 & 51.61 & 56.99 & 27.24 & 26.88 \\ 
Mathstral-7b-v0.1 & 36.56 & 36.20 & 48.39 & 14.70 & 16.49 \\ 
NuminaMath-7B-CoT & 43.73 & 40.14 & 47.31 & 17.20 & 17.20 \\ 
MetaMath-13B-V1.0 & 21.15 & 7.53 & 11.11 & 5.73 & 3.58 \\ 
MAmmoTH2-8B & 12.90 & 17.92 & 31.18 & 7.53 & 5.73 \\ 
\bottomrule
\end{tabular}
}
\label{tab:OIC}
\end{table*}

\begin{table*}[htbp]
\caption{Effects of in-context learning (ICL) with original example on \HARD. The percentages of $n(\text{correct} \to \text{wrong})$ are normalized by the number of errors with ICL, while the percentages of $n(\text{wrong} \to \text{correct})$ are by the number of errors without ICL. }
\centering
\resizebox{\textwidth}{!}{
\begin{tabular}{lcccc}
 \toprule
Model & num. errors (zero-shot) & num. errors (ICL w. original) & $n(\text{correct} \to \text{wrong})$ & $n(\text{wrong} \to \text{correct})$ \\ 
\midrule
Gemini-2.0-flash-thinking-exp & 61 (21.86 \%) & 58 (20.79 \%) & 17 (29.31 \%) & 20 (32.79 \%) \\ 
o1-preview & 77 (27.60 \%) & 72 (25.81 \%) & 21 (29.17 \%) & 26 (33.77 \%) \\ 
o1-mini & 60 (21.51 \%) & 60 (21.51 \%) & 24 (40.00 \%) & 24 (40.00 \%) \\ 
\midrule
Gemini-2.0-flash-exp & 92 (32.97 \%) & 91 (32.62 \%) & 30 (32.97 \%) & 31 (33.70 \%) \\ 
Gemini-1.5-pro & 121 (43.37 \%) & 110 (39.43 \%) & 27 (24.55 \%) & 38 (31.40 \%) \\ 
GPT-4o & 169 (60.57 \%) & 159 (56.99 \%) & 31 (19.50 \%) & 41 (24.26 \%) \\ 
GPT-4-turbo & 183 (65.59 \%) & 170 (60.93 \%) & 33 (19.41 \%) & 46 (25.14 \%) \\ 
Claude-3.5-Sonnet & 171 (61.29 \%) & 141 (50.54 \%) & 27 (19.15 \%) & 57 (33.33 \%) \\ 
Claude-3-Opus & 205 (73.48 \%) & 186 (66.67 \%) & 35 (18.82 \%) & 54 (26.34 \%) \\ 
\midrule
Llama-3.1-8B-Instruct & 251 (89.96 \%) & 249 (89.25 \%) & 18 (7.23 \%) & 20 (7.97 \%) \\ 
Gemma-2-9b-it & 246 (88.17 \%) & 239 (85.66 \%) & 14 (5.86 \%) & 21 (8.54 \%) \\ 
Phi-3.5-mini-instruct & 239 (85.66 \%) & 239 (85.66 \%) & 17 (7.11 \%) & 17 (7.11 \%) \\ 
\midrule
Deepseek-math-7b-rl & 241 (86.38 \%) & 236 (84.59 \%) & 19 (8.05 \%) & 24 (9.96 \%) \\ 
Qwen2.5-Math-7B-Instruct & 203 (72.76 \%) & 204 (73.12 \%) & 32 (15.69 \%) & 31 (15.27 \%) \\ 
Mathstral-7b-v0.1 & 238 (85.30 \%) & 233 (83.51 \%) & 19 (8.15 \%) & 24 (10.08 \%) \\ 
NuminaMath-7B-CoT & 231 (82.80 \%) & 231 (82.80 \%) & 23 (9.96 \%) & 23 (9.96 \%) \\ 
MetaMath-13B-V1.0 & 263 (94.27 \%) & 269 (96.42 \%) & 11 (4.09 \%) & 5 (1.90 \%) \\ 
MAmmoTH2-8B & 258 (92.47 \%) & 263 (94.27 \%) & 12 (4.56 \%) & 7 (2.71 \%) \\
\bottomrule
\end{tabular}
}
\label{tab:icl:breakdown}
\end{table*}

\clearpage
\subsection{Ablation Study: In-Context Learning with the Original Example v.s. In-Context Learning with a Random Example}

In \cref{tab:oic_sic}, we compare (1) the performance of one-shot in-context learning with the corresponding \textbf{original} unmodified (problem, solution) with (2) the performance of ICL with a \textbf{random} problem and solution chosen from the same category as the query problem. We find that ICL with the \textbf{original} problem and solution consistently outperforms ICL with a \textbf{random} example except for only one case.

\begin{table*}[htbp]
\vspace{-3mm}
\caption{Performance comparisons without and with the original problem and solution as the in-context learning example.}
\centering
\resizebox{0.75\textwidth}{!}{
\begin{tabular}{lcccc}
 \toprule
\multirow{2}{*}{\textbf{Model}} & \multicolumn{2}{c}{\textbf{\SAME}} & \multicolumn{2}{c}{\textbf{\HARD}}   \\ 
\cmidrule(r){2-3}  \cmidrule(r){4-5}
& ICL w. original & ICL (random) & ICL w. original  & ICL (random)\\ 
\midrule
o1-mini & \textbf{94.98} & 92.83 & \textbf{78.49} & 75.99 \\ 
\midrule
Gemini-1.5-pro & \textbf{88.17} & 75.99 & \textbf{60.57} & 51.97 \\ 
GPT-4o & \textbf{77.06} & 63.08 & \textbf{43.01} & 37.28 \\ 
GPT-4-turbo & \textbf{69.89} & 57.71 &\textbf{ 39.07} & 32.62 \\ 
Claude-3.5-Sonnet & \textbf{83.15} & 62.37 & \textbf{49.46} & 40.86 \\ 
Claude-3-Opus & \textbf{68.10} & 45.52 & \textbf{33.33} & 23.66 \\ 
\midrule
Llama-3.1-8B-Instruct & \textbf{36.56} & 28.32 & \textbf{10.75} & 6.45 \\ 
Gemma-2-9b-it & \textbf{42.65} & 27.60 & \textbf{14.34} & 12.90 \\ 
Phi-3.5-mini-instruct & \textbf{36.92} & 20.07 & \textbf{14.34} & 10.39 \\ 
\midrule
Deepseek-math-7b-rl & \textbf{45.52} & 34.41 & \textbf{15.41} & 13.26 \\ 
Qwen2.5-Math-7B-Instruct & \textbf{56.99} & 55.20 & \textbf{26.88} & 26.16 \\ 
Mathstral-7b-v0.1 & \textbf{48.39} & 24.37 & \textbf{16.49} & 8.96 \\ 
NuminaMath-7B-CoT & \textbf{47.31 }& 24.73 & \textbf{17.20} & 10.04 \\ 
MetaMath-13B-V1.0 & \textbf{11.11} & 8.60 & 3.58 & \textbf{5.38} \\ 
MAmmoTH2-8B & \textbf{31.18} & 3.94 &\textbf{ 5.73 }& 2.15 \\ 
\bottomrule
\end{tabular}
}
\label{tab:oic_sic}
\end{table*}

\subsection{Inference-time Scaling Behaviors}
\label{sec:inference:scaling}

\begin{figure*}[ht]
    \centering
    \includegraphics[width=0.32\linewidth]{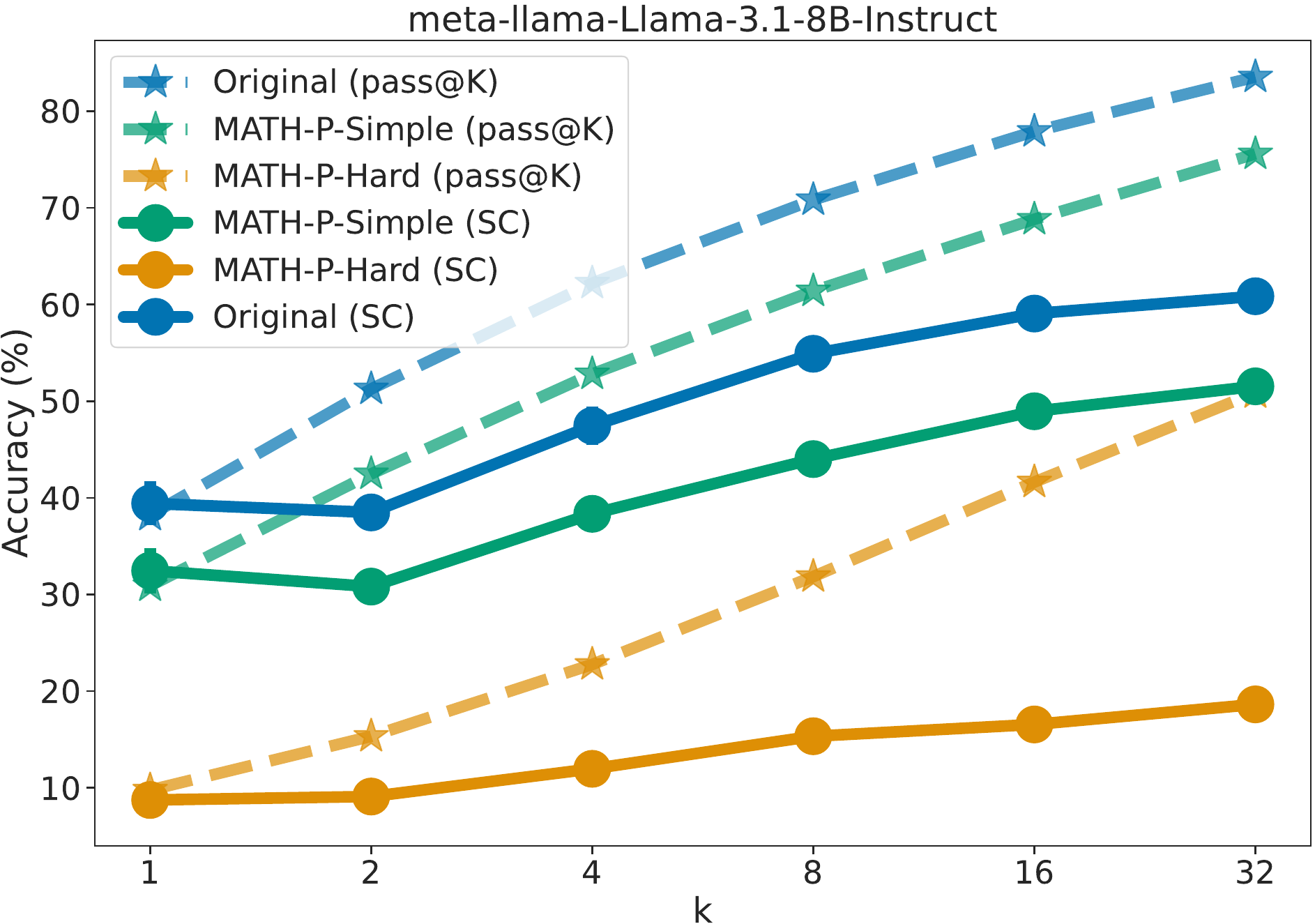}
    \includegraphics[width=0.32\linewidth]{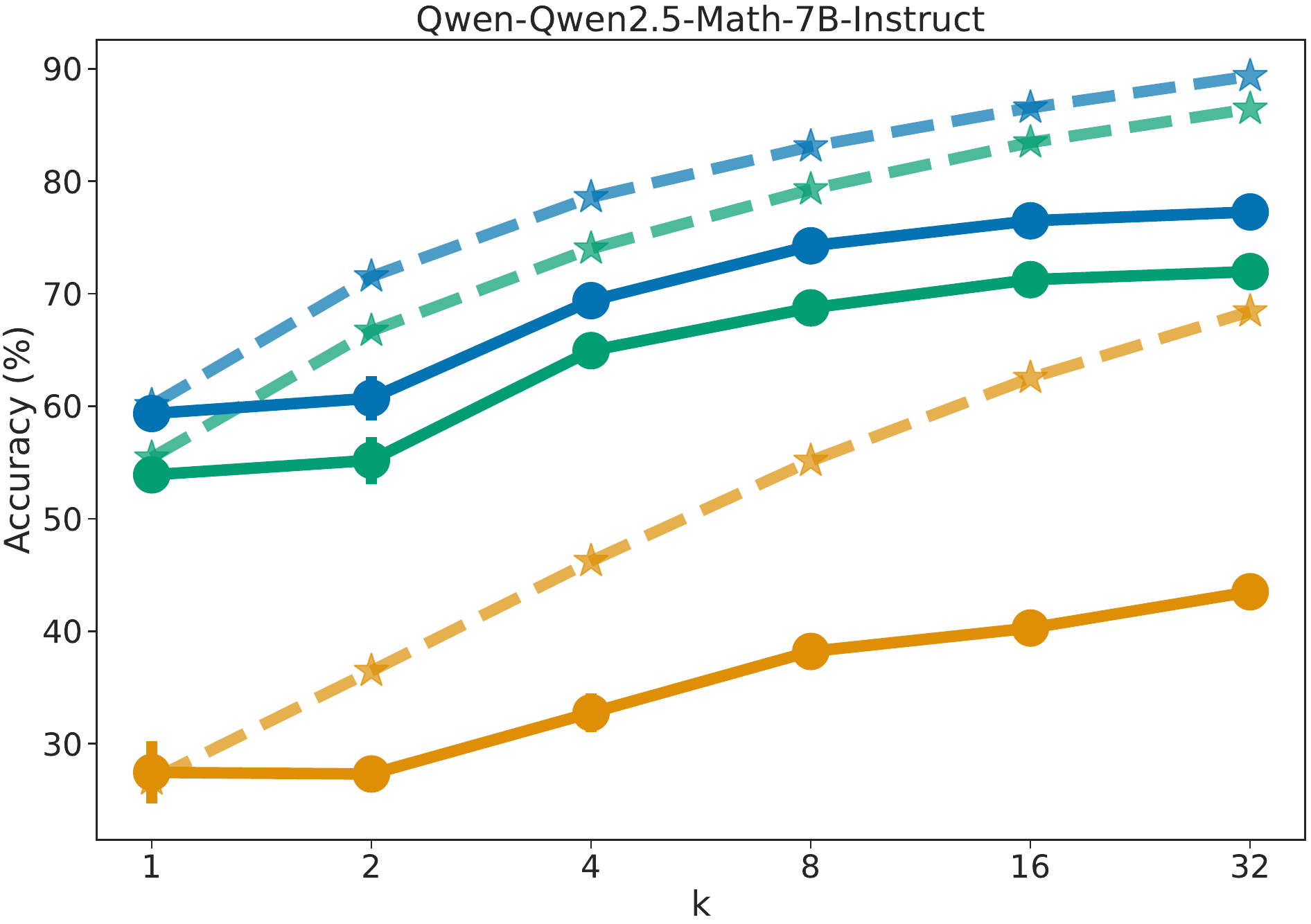}
    \includegraphics[width=0.32\linewidth]{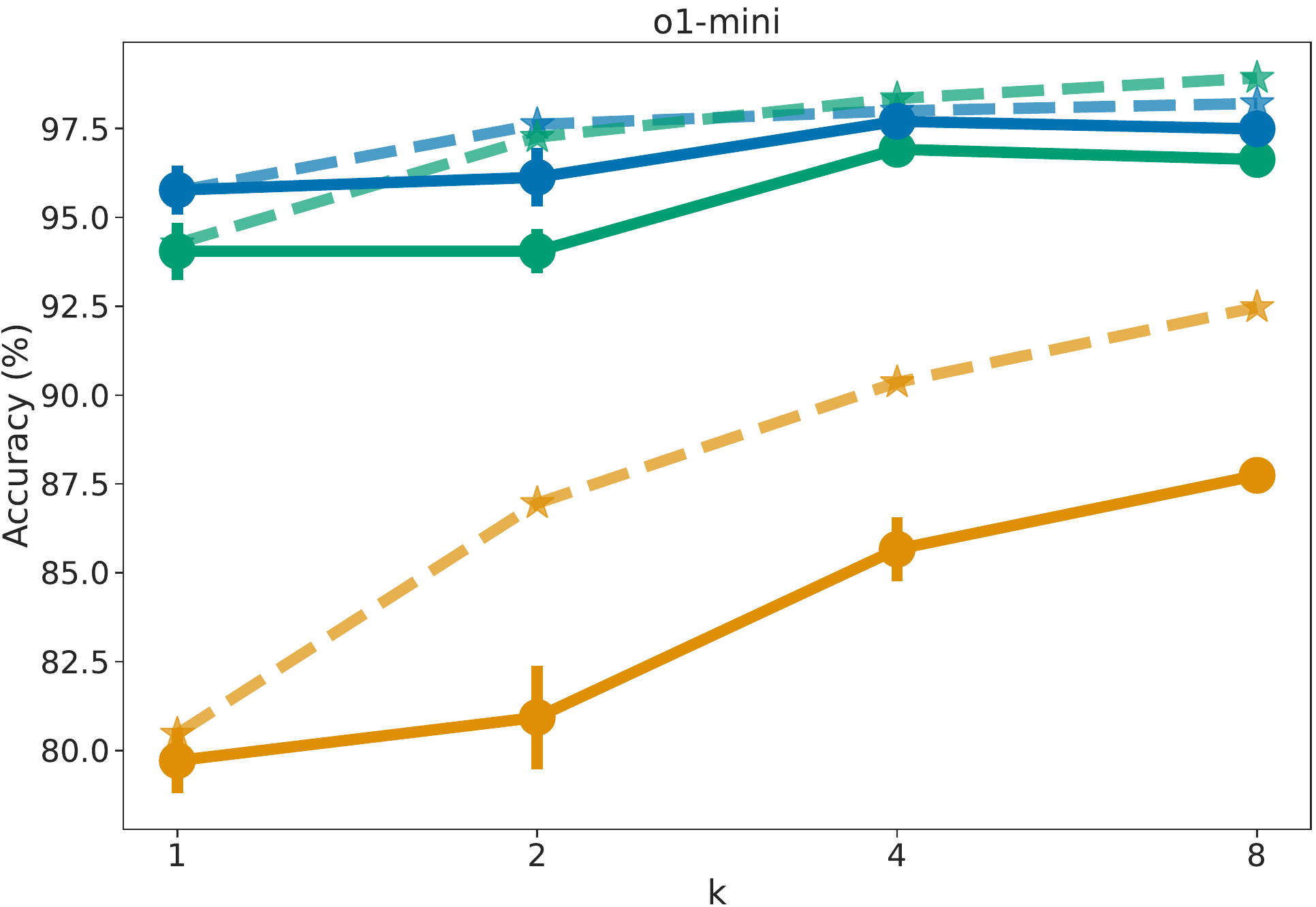}
    \caption{The effect of scaling up inference-time compute. We report pass@k and self-consistency (SC) accuracies for different numbers of solutions $k$.
    }
    \label{fig:inference:scaling}
\end{figure*}

In this subsection, we investigate the inference-time scaling behaviors of LLMs on our benchmarks. 
We compute the pass@k metric following~\citet{chen2021codex}. Specifically, for each problem, we generate $N$ solutions independently, and compute the pass@k metric via the following formula for each $1\leq k\leq N$:
\[
    \mathrm{pass@k} = \mathbb{E}_{\mathrm{problem}} \left[ 1-\frac{ { N-c \choose k}}{ {N \choose k} } \right], \text{ where } c \text{ is the number of correct answers of the } n \text{ runs}.
\]
We also compute the performance of self-consistency~\citep{wang2022self}, a.k.a., majority voting, where for each $k$, we randomly sample $k$ responses from the $N$ runs and get the majority-voted answer. We report the average and standard deviation among 5 random draws.
We only evaluate three models: o1-mini, Llama-3.1-8B-Instruct, and Qwen2.5-Math-7B-Instruct. For Llama-3.1-8B-Instruct, and Qwen2.5-Math-7B-Instruct, we choose $N=64$, while for o1-mini we set $N=8$. The results are plotted in \cref{fig:inference:scaling}.

\end{document}